\documentclass[10pt,twocolumn,letterpaper]{article}

\usepackage{wacv}
\usepackage{times}
\usepackage{epsfig}
\usepackage{graphicx}
\usepackage{amsmath}
\usepackage{amssymb}
\usepackage{url}
\usepackage{booktabs} 
\usepackage{multirow}
\usepackage{commath}
\usepackage{amsmath}
\usepackage{dblfloatfix}
\usepackage{caption}
\usepackage{multicol}
\usepackage{makecell}
\usepackage{rotating}
\usepackage{subcaption}
\usepackage{enumitem}

\usepackage[utf8]{inputenc}  
\usepackage[T1]{fontenc}   

\makeatletter
\@addtoreset{section}{part}
\makeatother

\usepackage{widetext}

\wacvfinalcopy 

\def\wacvPaperID{653} 

\ifwacvfinal\fi
\begin{document}

\title{ScaIL: Classifier Weights Scaling for Class Incremental Learning}

\author{Eden Belouadah, Adrian Popescu\\
Université Paris-Saclay, CEA, Département Intelligence Ambiante et Systèmes Interactifs,\\
91191 Gif-sur-Yvette, France\\
{\tt\small {eden.belouadah,adrian.popescu}@cea.fr}
}

\maketitle
\ifwacvfinal\thispagestyle{empty}\fi

\begin{abstract}
Incremental learning is useful if an AI agent needs to integrate data from a stream. 
The problem is non trivial if the agent runs on a limited computational budget and has a bounded memory of past data.
In a deep learning approach, the constant computational budget requires the use of a fixed architecture for all incremental states.
The bounded memory generates imbalance in favor of new classes and a prediction bias toward them appears. 
This bias is commonly countered by introducing a data balancing step in addition to the basic network training.
We depart from this approach and propose simple but efficient scaling of past classifiers' weights to make them more comparable to those of new classes.
Scaling exploits incremental state statistics and is applied to the classifiers learned in the initial state of classes to profit from all their available data. 
We also question the utility of the widely used distillation loss component of incremental learning algorithms by comparing it to  vanilla fine tuning in presence of a bounded memory. 
Evaluation is done against competitive baselines using four public datasets.
Results show that the classifier weights scaling and the removal of the distillation are both beneficial.

\end{abstract}

\vspace{-1em}
\section{Introduction}
Artificial agents are often deployed in dynamic environments in which information often arrives in streams~\cite{DBLP:journals/corr/abs-1802-07569}. 
For instance, this is the case of a robot which operates in an evolving environment or of a face recognition app which needs to deal with new identities.
In such settings, an incremental learning (IL) algorithm is needed to increase the recognition capacity when integrating new data.  
There was recently a strong regain of interest for IL with the adaptation of deep learning methods~\cite{DBLP:conf/cvpr/AljundiCT17, DBLP:conf/eccv/CastroMGSA18,DBLP:conf/cvpr/RebuffiKSL17,DBLP:journals/corr/RusuRDSKKPH16}.
Incremental learning is non-trivial if the artificial agents have limited computational and memory budgets.
If the memory of past classes is bounded or unavailable, the system underfits past data when new information is integrated and catastrophic forgetting~\cite{mccloskey:catastrophic} occurs. 
Since a joint optimization of computational and memory requirements is hard, if not impossible, existing IL algorithms focus on one of these two aspects. 
In a first scenario~\cite{DBLP:conf/cvpr/AljundiCT17,DBLP:conf/cvpr/MallyaL18,DBLP:journals/corr/RusuRDSKKPH16,DBLP:conf/cvpr/WangRH17}, the number of deep model parameters is allowed to grow and no memory is used. 
In a second scenario~\cite{DBLP:conf/eccv/CastroMGSA18,DBLP:conf/bmvc/He0SC18,DBLP:journals/corr/abs-1807-02802,DBLP:conf/cvpr/RebuffiKSL17,DBLP:journals/corr/abs-1904-01769}, the deep architecture is fixed and a memory is introduced for past class exemplars to alleviate the effect of catastrophic forgetting. 
These algorithms update deep models by adapting the fine tuning procedure to include classification and distillation losses. 

\begin{figure*}[t]
	\begin{center}
\includegraphics[width=0.7\textwidth,trim={0cm 8cm 0cm 3.3cm}]{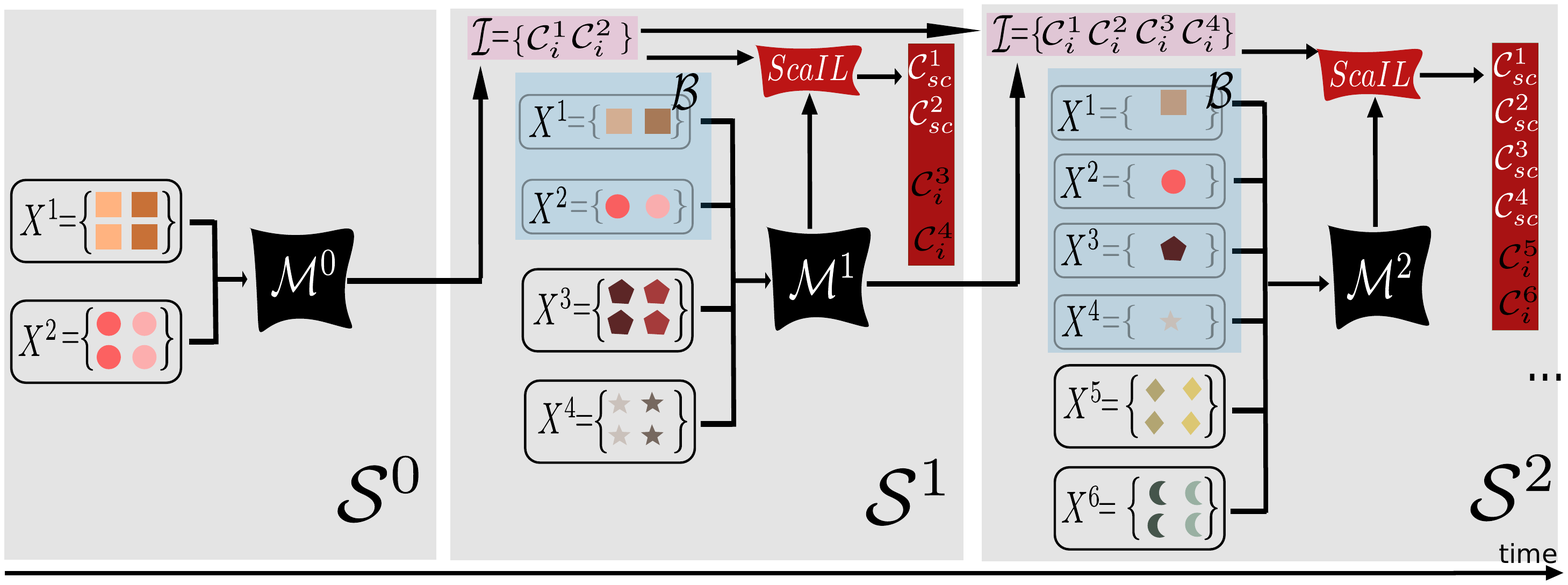}
	\end{center}
	\caption{Illustration of $ScaIL$.
	States are noted $\mathcal{S}^k$, image data $X^j$, deep models $\mathcal{M}^k$ and classifier weights $C^{j}_{i}$ and $C^{j}_{sc}$, where: $j$ is the class label, $i$ is the initial state in which the classifier was learned with all data and $sc$ means that the classifier was scaled using $ScaIL$. 
	We represent three states $\mathcal{S}^0$, $\mathcal{S}^1$ and $\mathcal{S}^2$ which recognize $2, 4$ and  $6$ classes respectively.
	The bounded memory (light blue), is fixed at $\mathcal{B}=4$ past classes exemplars. 
	As the training advances, the data imbalance between past and new classes grows due to bounded $\mathcal{B}$ and the prediction bias in favor of new classes becomes more prominent.
	$ScaIL$ reduces this bias by making classifier weights of past and new classes more comparable by using a small memory $\mathcal{I}$ which stores initial classifiers $C^{j}_{i}$. 
	In each IL state, $ScaIL$ replaces the raw classifiers of past classes provided by the model $\mathcal{M}^k$ by $C_{sc}^j$, a scaled version of $C_i^j$, the initial classifier. 
	Since $ScaIL$ combines classifiers learned in different IL states, initial classifiers are reshaped using aggregate statistics from the current and the initial states. The classifiers for newly learned classes are left as learned by the current model $\mathcal{M}^k$. \textit{Best viewed in color.}
    }
      \vspace{-1em}
	\label{fig:overview}
\end{figure*}

We focus on this second scenario and introduce $ScaIL$, a method which reduces the bias in favor of new classes by exploiting the classifier weights of past classes as learned in their initial state with all class data available.
Since past class classifiers are learned in different previous IL states, they are reshaped to be usable in the current state. 
Their scaling uses aggregate statistics from the current and initial states. $ScaIL$ is illustrated in Figure~\ref{fig:overview} with a toy example which includes an initial and two incremental states.
In addition to the bounded exemplar memory $\mathcal{B}$,
$ScaIL$ requires the use of a compact memory $\mathcal{I}$ which stores the classifier weights from the initial states of past classes.
A second contribution, of practical importance, is to simplify the deep model update across incremental states.
The widely used distillation loss term~\cite{DBLP:conf/eccv/CastroMGSA18,DBLP:conf/bmvc/He0SC18,DBLP:journals/corr/abs-1807-02802,DBLP:conf/cvpr/RebuffiKSL17,DBLP:journals/corr/abs-1904-01769} is ablated here and model updates are done with vanilla fine tuning.

Evaluation is done with four public datasets and three values for the number of incremental states $\mathcal{Z}$ and the exemplar memory $\mathcal{B}$, the two key components of class IL algorithms.
$ScaIL$ is compared to strong baselines from literature and to new ones proposed here and the obtained results indicate that it has the best overall performance. 

\section{Related Works}
Incremental learning is an open research topic which recently witnessed a regain of interest with the use of deep learning algorithms. 
We discuss three groups of methods which focus on different parts of IL. 
Due to limited space, only a representative subset of methods is described.

A first group increases the number of parameters of deep architectures to accommodate new classes. Growing a Brain~\cite{DBLP:conf/cvpr/WangRH17} increases the depth and/or the width of the layers to integrate new classes. In Progressive Neural Networks~\cite{DBLP:journals/corr/RusuRDSKKPH16}, a new network is added for each new task and lateral connections are used between all networks to share the representation. 
$PackNet$~\cite{DBLP:conf/cvpr/MallyaL18} uses weight pruning techniques to free up redundant network parameters.
When new classes arrive, the freed up parameters are attributed to the new task. 
The number of parameters grows slowly but only a limited number of new tasks can be added. 
These algorithms are a good choice if the complexity of deep models can grow across incremental states. 
However, they increase the model memory footprint and slow down inference, especially for a large number of incremental states.

A second, less frequent group, is based on fixed representations.
Here, the feature extractor network does not evolve across IL states. 
$FearNet$~\cite{DBLP:conf/iclr/KemkerK18} is biologically inspired by the functioning of human brain. 
The incremental learning process is implemented with three networks which model short and long term memory and a decision network to choose the activated network.
The main drawback of $FearNet$ is that memory grows in a nearly linear fashion across IL states because detailed statistics about past classes are needed.
$DeeSIL$~\cite{deesil} is an adaptation of transfer learning to a class IL context~\cite{DBLP:journals/corr/abs-1805-08974,DBLP:conf/cvpr/KuzborskijOC13}.
It learns a fixed representation on the first state and deploys a set of SVMs to increment recognition capacity afterwards.
The reported top-5 accuracy on
ILSVRC~\cite{DBLP:journals/ijcv/RussakovskyDSKS15} with a bounded memory $\mathcal{B}=20000$ exemplars is 74.7\%.
A fixed representation is tested in~\cite{DBLP:conf/nips/RebuffiBV17} but its past classes are unnecessarily relearned only with exemplars in each IL states and results are suboptimal.
The main advantage of fixed representations is that all positive examples can be used for all classes since the deep model does not evolve across incremental states.
However, performance depends heavily on the quality of the initial representation. 
If the representation is learned on small dataset or if the new classes are significantly different from the initial ones, the generalization capacity is low.

A third influential group of algorithms updates deep models across incremental states using an adapted fine tuning procedure.
These algorithms are inspired by Learning-without-Forgetting ($LwF$)~\cite{DBLP:conf/eccv/LiH16}, which introduces a distillation loss term to handle catastrophic forgetting in absence of a memory of past classes. 
This term encourages the network to reproduce the same outputs for past classes in the current state as in past ones.
We discuss some representative adaptations of distillation to IL hereafter.
$iCaRL$ ~\cite{DBLP:conf/cvpr/RebuffiKSL17} implements $LwF$ using binary cross-entropy loss, which operates independently on class outputs.
The use of this loss is not clearly justified but it is assumed to cope with class imbalance~\cite{DBLP:conf/iccv/LinGGHD17}. 
Also, the authors modify the distillation term to use sigmoids instead of the standard softened softmax targets. 
$iCaRL$ adds the following steps for an efficient adaptation to an IL context: (1) exploit a bounded memory for past classes, (2) select exemplars using a herding mechanism which approximates the real class mean ~\cite{DBLP:conf/icml/Welling09} and (3) replace the outputs of the deep models by a
Nearest-Exemplars-Mean (NEM) external classifier, an adaptation of nearest-class-mean~\cite{DBLP:journals/pami/MensinkVPC13}, to tackle class imbalance. 
$iCaRL$ top-5 accuracy reaches 62.5\% on ILSVRC~\cite{DBLP:journals/ijcv/RussakovskyDSKS15} dataset with a memory of $\mathcal{B}=20000$ exemplars. 
The authors also conclude that vanilla fine tuning is not fitted for IL with bounded memory. 
However, their main experiment tests $iCaRL$ with memory and vanilla fine tuning without memory and the comparison is not fair.
In an additional experiment, they compare the two methods only on a small scale dataset and, while $iCaRL$ remains superior, the gap between the two methods is much smaller.
End-to-end incremental learning~\cite{DBLP:conf/eccv/CastroMGSA18} uses a distillation component which is closer to the original definition from~\cite{DBLP:journals/corr/HintonVD15}.  
The authors exploit standard cross-entropy loss and their basic distilled network has performance similar to that of $iCaRL$.
A balanced fine tuning step is added to tackle data imbalance and data augmentation is also used.
As a result, the method gains 7 points over $iCaRL$ on ILSVRC with $\mathcal{B}=20000$.
In~\cite{DBLP:journals/corr/abs-1807-02802}, the authors show that the use of higher temperature to soften distillation helps to some extent.
Very recently, the authors of~\cite{DBLP:journals/corr/abs-1904-01769} introduced a multi-model and multi-level distillation for IL.
The method incorporates knowledge from all previous incremental states and not only from the latest one. 
A performance improvement of 3 to 5 points over $iCaRL$  is reported. 
Generative Adversarial Networks were also considered as a mean to generate image exemplars for past classes instead of storing them directly~\cite{DBLP:conf/bmvc/He0SC18}.
While the approach is appealing, the quality of generated images is still insufficient. 
A combination of generated and real images was necessary to slightly enhance performance over $iCaRL$. 
$BiC$~\cite{DBLP:conf/cvpr/WuCWYLGF19} is a very recent approach that handles catastrophic forgetting by adding a linear model after the last fully connected layer to correct the bias towards new classes. We add $BiC$ to the results table for SotA completeness.
Approaches from this group tend to cope well with the integration of new data but retraining the network at each incremental step is costly. 

\section{Proposed Method}
\subsection{Class IL Problem Formalization}
We focus on IL with constant model complexity, $\mathcal{Z}$ incremental states and a bounded memory $\mathcal{B}$ of past classes. The proposed formalization is adapted from~\cite{DBLP:conf/eccv/CastroMGSA18,DBLP:conf/bmvc/He0SC18,DBLP:conf/cvpr/RebuffiKSL17}.
We note: $\mathcal{S}^k$ - the incremental state, $N_k$ - the number of classes in $\mathcal{S}^k$, $\mathcal{X}^{N_k}$ - the training dataset in $\mathcal{S}^k$, $\mathcal{M}^k$ -  the deep model and $\mathcal{C}^{N_k}$ - the classifier weights layer.
The initial state $\mathcal{S}^0$ includes a dataset $\mathcal{X}^{N_0}=\{X^1, X^2, ..., X^{N_0}\}$ with $N_0 = P_0$ classes. $X^j=\{x_1^j,x_2^j, ..., x_{n_j}^j\}$ is the set of $n_j$ training examples for the $j^{th}$ class. 
An initial model $\mathcal{M}^0:\mathcal{X}^{N_0} \rightarrow \mathcal{C}^{N_0}$ is trained to recognize $N_0$ classes using all data from $\mathcal{X}^{N_0}$.
$P_k$ new classes need to be integrated in each incremental state $\mathcal{S}^k$, with $k>0$. Each IL step updates the previous model $\mathcal{M}^{k-1}$ into the current model $\mathcal{M}^k$ which recognizes $N_k = P_0 + P_1 + ... + P_k$ classes in incremental state $\mathcal{S}^k$.
All data of the $P_k$ new classes are available but only a bounded exemplar memory $\mathcal{B}$ of the $N_{k-1}$ past classes  is allowed.
If memory allocation is balanced, each past class is represented by $\frac{\mathcal{B}}{N_{k-1}}$ exemplars.
We note $\mathcal{M}^k:\mathcal{X}^{N_k} \rightarrow \mathcal{C}^{N_k}$ the model which transforms the $\mathcal{X}^{N_k}$ dataset into a set of raw classifiers $\mathcal{C}^{N_k} = \{C_k^1, C_k^2,..., C_k^{N_{k-1}}, C_k^{N_{k-1}+1},..., C_k^{N_k}\}$.
The classifier weights learned in state $\mathcal{S}^k$ for the $j^{th}$ class are written $C_k^j = \{w^1(C_k^j), w^2(C_k^j),...,w^D(C_k^j)\}$, where $D$ is the size of the features extracted from the penultimate layer of $\mathcal{M}^{k}$. 

\subsection{Classifier Weights Scaling}
\begin{figure*}[t]
	\begin{center}
\includegraphics[width=0.7\textwidth,trim={1cm 7cm 2.5cm 1cm}]{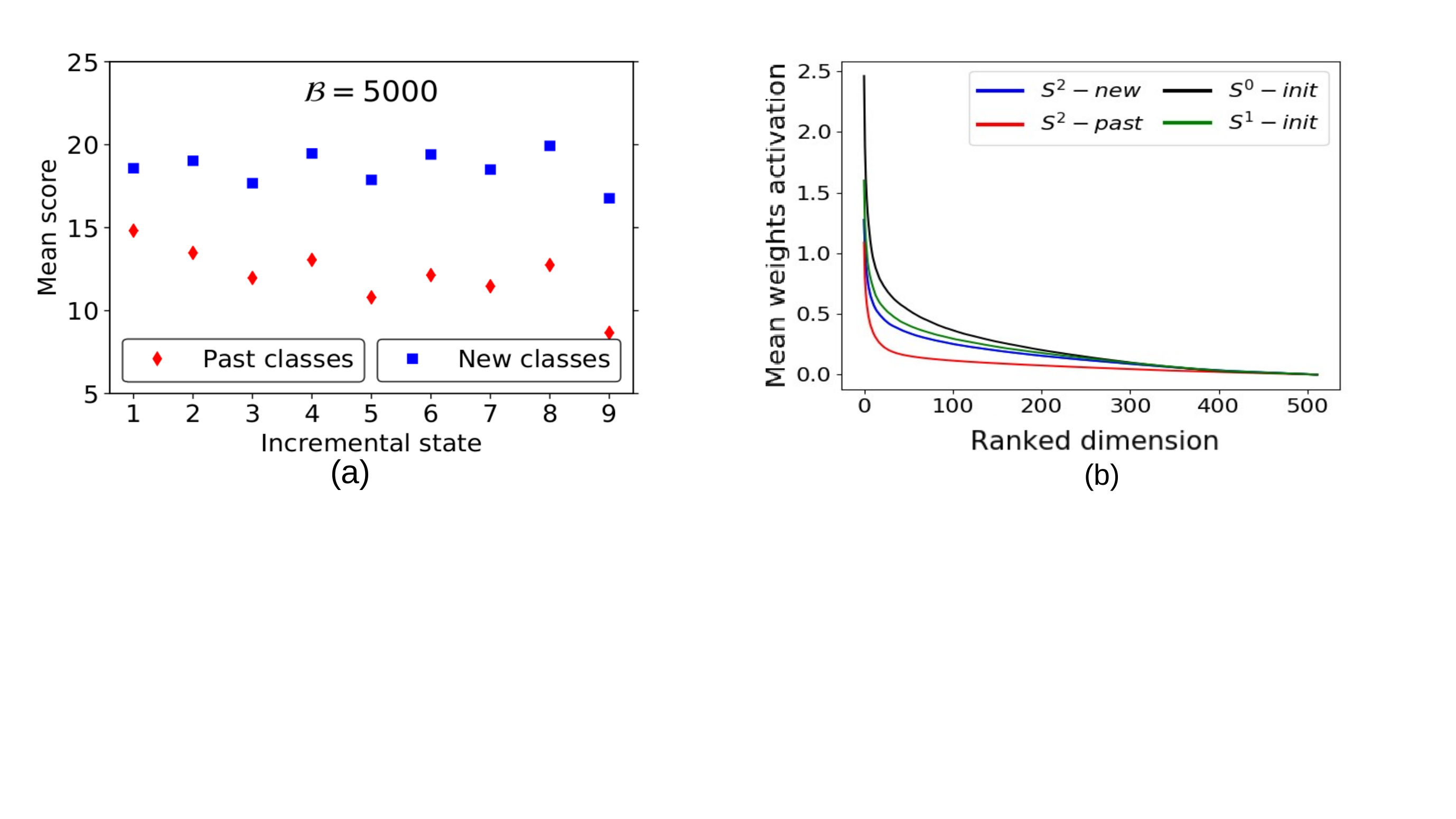}
\end{center}
\caption{(a) - Raw prediction scores (before softmax) of vanilla fine tuning for the ILSVRC dataset~ with $\mathcal{B}=5000$ past exemplars and a total of $\mathcal{Z}=10$ states. Incremental states from 1 to 9 are represented. The initial state $\mathcal{S}^0$ is non-incremental and is not shown. (b) - Ranked mean weight activations of new and past classes in state $\mathcal{S}^2$ (blue and red) and mean weight activations of $\mathcal{S}^2$ past classes as initially learned in $\mathcal{S}^0$ (black) and $\mathcal{S}^1$ (green). \textit{Best viewed in color.}
    }
\label{fig:scores}
\end{figure*}

Incremental learning algorithms strive to approach the performance of full learning, in which the entire training set is available for all classes at all times. 
When a bounded set of past exemplars is stored, a prediction bias toward new classes appears due to the data imbalance in their favor.
This bias is illustrated in Figure~\ref{fig:scores}(a) with the difference between mean raw predictions for past and new classes after incrementally fine tuning the $ILSVRC$ dataset~\cite{DBLP:journals/ijcv/RussakovskyDSKS15} with $\mathcal{B}=5000$ past exemplars.
The average score difference in favor of new classes over all incremental states is 6.45 points. 
The prediction gap is due to the stronger activations of classifier weights for new classes compared to past classes, as illustrated by the blue and red curves from Figure~\ref{fig:scores}(b).
It is thus tempting to try to reshape the classification layers of past and new classes in order to make them more comparable. 
A simple way to do this is to add a normalization layer to the current deep model and we provide results with such a baseline ($FT^{L2}$) in Section~\ref{sec:expe}.

$ScaIL$ attempts to approximate full learning by exploiting past classifiers as learned in their initial state, with all images available. 
Since the deep models evolve during the incremental process, a transformation of the initial classifiers is needed for them to be usable in the current incremental state. 
The method is illustrated in Figure~\ref{fig:overview}.

The main differences with existing IL algorithms which exploit a bounded memory are: (1) the introduction of a second memory $\mathcal{I}$ to store initial past class classifiers and (2) the ablation of the distillation loss.
Note that the size of $\mathcal{I}$ is orders of magnitude smaller than that of $\mathcal{B}$ since it only stores hundreds of floating point values per class instead of exemplar images. 
The immediate advantage of the method is that initial classifiers of past data are learned with all data.
Initial classifiers learned with all images are stronger than the past classifiers learned only with exemplars in the current state. 
This is clearly visible in Figure~\ref{fig:scores}(b) from the comparison of past classifiers weights as learned in the current state (red) and the weights of the same classifiers learned in states $\mathcal{S}^0$ and $\mathcal{S}^1$ (black and green).
We also note the activations of new classes become weaker as the incremental learning process advances.
The new classes from state $\mathcal{S}^0$ (black) are the strongest, followed by new classes from $\mathcal{S}^1$ (green) and those from $\mathcal{S}^2$ (blue). 

The main challenge associated to $ScaIL$ is to combine classifiers originating from deep models learned in different IL states.
The reuse of initial classifiers in later incremental states is made possible by fine tuning process with a memory of the past.
This process results in a partial preservation of the feature space even if the deep model evolves.
In the supplementary material, we show that classifier reuse across states is impossible in absence of memory during IL model updates.
$ScaIL$ reshapes initial classifiers from $\mathcal{I}$ in order to make them comparable to those of newly learned classes in the feature space defined by the current state's deep model. 
The scaling is based on weights statistics computed for initial models in each incremental state (Equation~\ref{eq:meanpast}).
Before computing the means in the equation, the weights of each initial classifier are ranked by their absolute value. 
The use of absolute values is necessary since classifier weights activations can be positive or negative.


\begin{equation}
    \mu_{i}^{rank} = \frac{1}{P_i} \times \sum_{j=1}^{P_i} \lvert w^{rank}(C_i^j)\rvert
    \label{eq:meanpast}
\end{equation}

$\mu_{i}^{rank}$ is the mean of the weights ranked $rank$, with $1 \leq rank \leq D $, for the $P_i$ classes initially learned in each past state $\mathcal{S}^i$, with $0 \leq i < k$.
Figure~\ref{fig:scores}(b) shows that classifiers of each past state have different statistical distributions.
To make class predictions from  different states comparable, it is necessary to compute $\mu_{i}^{rank}$ separately for each state.
If $k=i$, we compute $\mu_{k}^{rank}$, the mean of classifier weights for new classes from the current state $\mathcal{S}^k$, which is  also their initial state. 
Note that each mean is computed using weights situated at the same rank for each classifier. 
For instance, $\mu_{k}^{1}$ and $\mu_{k}^{D}$ will aggregate respectively the maximum and minimum weights of newly learned classes in $\mathcal{S}^k$.

$ScaIL$ transforms the past classifier weights as learned in their initial state using Equation~\ref{eq:scail}.
$w_{sc}^h(C_{sc}^j)$ is the scaled version of $w^h(C_i^j)$, the $h^{th}$ dimension of the initial classifier $C_i^j$ of the $j^{th}$ past class. 
These weights are scaled using the ratio between the mean activation of new classes and that of past classes in their initial state. 
In Equation~\ref{eq:scail}, each weight $w^h$ is scaled using the mean activations of its corresponding rank, returned by function $r(\cdot)$, in the current and initial states $\mathcal{S}^k$ and $\mathcal{S}^i$. 
For instance, if the first weight ($h=1$) of the classifier $C_i^j$ is ranked $9^{th}$, it will be scaled using the mean activations to the ninth dimension of the mean ranked activations $\mu_{k}^{9}$ and $\mu_{i}^{9}$ respectively.
This is done in order to preserve the relative importance of each classifier weight. 
Figure~\ref{fig:scores}(b) shows that $\mu_{i}^{rank} > \mu_{k}^{rank}$ for a given rank $rank$. 
Consequently, $ScaIL$ scaling reduces the weights of the $j^{th}$ class learned in its initial state to make it more comparable to classifiers of new classes from the current state. 
The scaled classifier for each past class of the current state $\mathcal{S}^k$ is written as $C_{sc}^j = \{w_{sc}^1(C_{sc}^{j}), w_{sc}^2(C_{sc}^{j}), ..., w_{sc}^D(C_{sc}^{j})\}$.
The $ScaIL$ classification layer for $\mathcal{S}^k$ combines scaled classifiers for past classes and original classifiers for new classes. It can be written as $\mathcal{C}_{sc}^{N_k} = \{C_{sc}^1, ..., C_{sc}^{N_{k-1}}, C_k^{N_{k-1}+1}, ..., C_k^{N_k}\}$.
The features learned in $\mathcal{S}^k$ are fed into this scaled classification layer instead of the original one provided by $\mathcal{M}^k$. 

\begin{equation}
    w_{sc}^h(C_{sc}^j) = \frac{\mu_{k}^{r(h)}}{\mu_{i}^{r(h)}} \times w^h(C_i^j)
    \label{eq:scail}
\end{equation}

Note that only scores of the top-10 past classes are scaled as they code more information, the scores of the remaining past classes are set to zero. The choice of this value is experimental.   

\begin{figure}[ht!]
	\begin{center}
\includegraphics[width=0.4\textwidth,trim={0cm 1.2cm 0cm 0.2cm}]{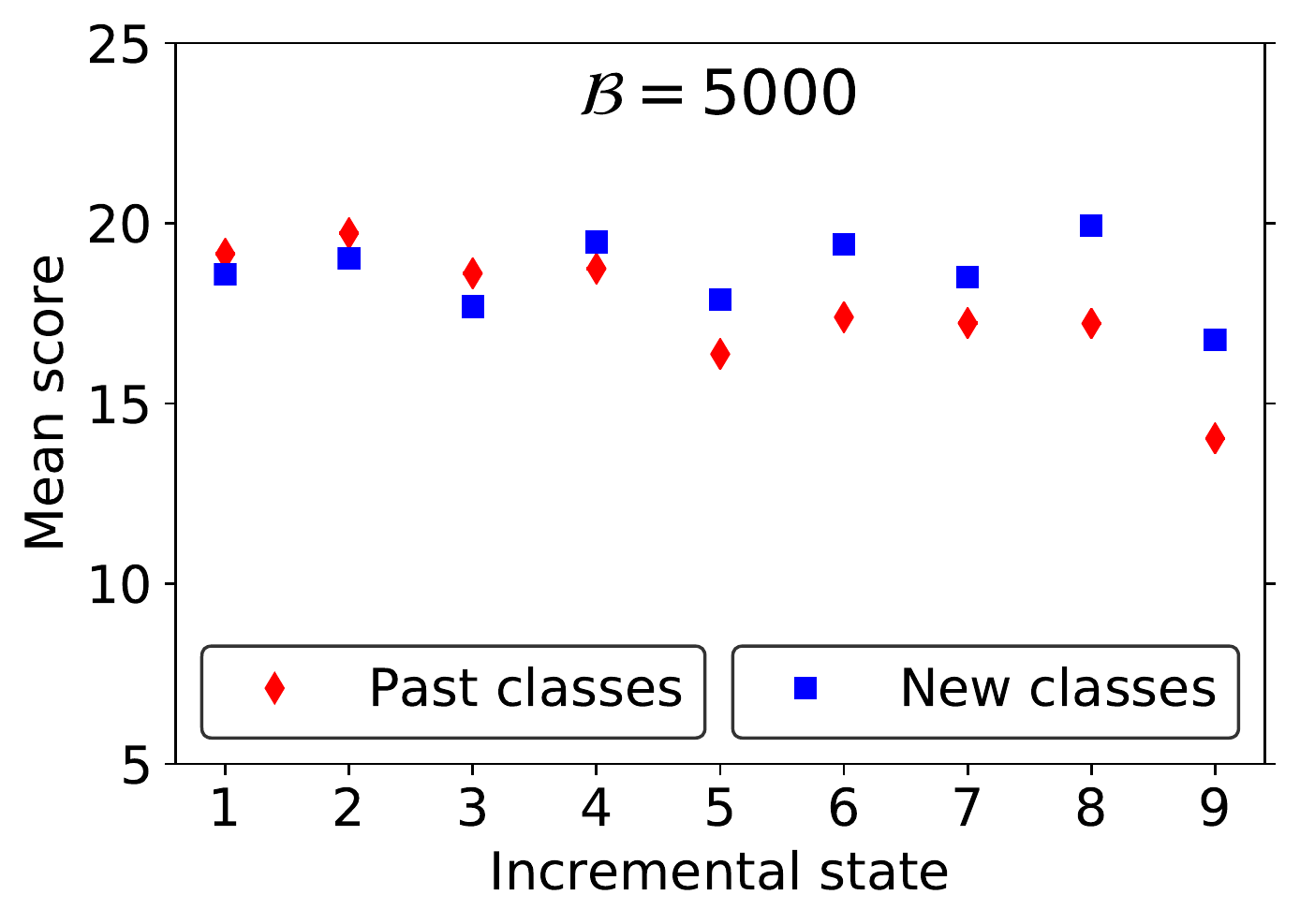}
	\end{center}
	\caption{Prediction scores after scaling for the ILSVRC dataset~\cite{DBLP:journals/ijcv/RussakovskyDSKS15} with $\mathcal{B}=5000$ exemplars and $\mathcal{Z}=10$ states.
    }
	\label{fig:scores_scail}
\end{figure}

We illustrate the effect of $ScaIL$ on the prediction scores in Figure~\ref{fig:scores_scail}. 
Past classes have a slightly larger mean classification score in the first states and a lower one in subsequent states. 
While not completely aligned, the predictions of past and new classes in $ScaIL$ are much more balanced compared to those of raw fine tuning results from Figure~\ref{fig:scores}(a).

\section{Experiments}
\label{sec:expe}
\subsection{Datasets}
Experiments are done with four public datasets.
This evaluation is more comprehensive than the usual one proposed in \cite{DBLP:conf/eccv/CastroMGSA18, DBLP:conf/cvpr/RebuffiKSL17}, which includes two datasets only. We use:
\begin{itemize}[leftmargin=*]
\item ILSVRC~\cite{DBLP:journals/ijcv/RussakovskyDSKS15}: a subset of ImageNet~\cite{DBLP:conf/cvpr/DengDSLL009} designed for object recognition and used in the popular ImageNet LSVRC challenges. 
For comparability, we retain the same configuration (order of classes and train/test splits) as~\cite{DBLP:conf/cvpr/RebuffiKSL17}. 
This version of ILSVRC contains 1000 classes, with 1,23 million training and 50,000 test images.

\item VGGFace2~\cite{DBLP:conf/fgr/CaoSXPZ18}: face recognition dataset including over 9000 identities in its full version. 
Due to the heavy computation associated to IL, we select the 1000 identities with the largest number of training images. 
The resulting dataset has 491,746 training and 50,000 test images.
Face detection was done using MTCNN ~\cite{DBLP:journals/spl/ZhangZLQ16} which was applied to each image prior to training and test phases.

\item Google Landmarks~\cite{DBLP:conf/iccv/NohASWH17} (Landmarks below): landmark recognition dataset whose full version includes over 30000 classes.
We select the top 1000 classes and experiment with 374,367 training and 20,000 test images.

\item CIFAR-100~\cite{Krizhevsky09learningmultiple}: object recognition dataset including 100 classes, with 500 training and 100 test images each. 
\end{itemize}
\subsection{Methodology and Baselines}
The experimental setup used here is inspired from~\cite{ DBLP:conf/eccv/CastroMGSA18, DBLP:conf/cvpr/RebuffiKSL17}.
The size of bounded memory $\mathcal{B}$ and the number of incremental states $\mathcal{Z}$ are the two most important parameters in IL with memory.
We use three different values for each of them while fixing the value of the other parameter.
\vspace{0.3em}

\textbf{Memory Management.}
$\mathcal{B}$ size is varied to evaluate the robustness of the tested methods with memory availability. 
We fix the number of states $\mathcal{Z}=10$ and run experiments with a memory which amounts to approximately $2\%, 1\%, 0.5\%$ of the full training sets. 
Memory sizes are thus $\mathcal{B}=\{20000, 10000, 5000\}$ for ILSVRC, $\mathcal{B}=\{10000, 5000, 2500\}$ for VGGFace2, $\mathcal{B}=\{8000, 4000, 2000\}$ for Landmarks and $\mathcal{B}=\{1000, 500, 250\}$ for CIFAR-100.
Whenever a new incremental state is added, memory is updated by inserting exemplars of new classes and reducing exemplars of past classes in order to fit the maximum size. 
\vspace{0.3em}

\textbf{Incremental States.} 
The number of incremental states is the second key component of IL with memory and we evaluate its variation. 
We fix the memory to $\mathcal{B}=0.5\%$ and test with $\mathcal{Z}=\{20, 50\}$ in addition to $\mathcal{Z}=10$.
The lowest memory size was selected since it is the most interesting configuration when memory budget is smallest.
\vspace{0.3em}

\textbf{Exemplar selection.} A herding mechanism~\cite{DBLP:conf/icml/Welling09}, called Nearest-Exemplars-Mean (NEM) was introduced in $iCaRL$ for exemplar selection ~\cite{DBLP:conf/cvpr/RebuffiKSL17}.  $BiC$ uses the same herding mechanism. For this, we provide results for $ScaIL$ with and without herding. $ScaIL^{herd}$ is directly comparable with $iCaRL$ and $BiC$.
 \vspace{0.3em}
 
\textbf{Evaluation measures.}
To facilitate comparability, each configuration is evaluated with the widely used top-5 accuracy~\cite{DBLP:journals/ijcv/RussakovskyDSKS15}.
Each algorithm is tested in a large number of configurations and it is important to propose a summarized performance score. 
Inspired by works such as~\cite{DBLP:conf/nips/RebuffiBV17,tamaazousti2018universal}, we introduce a global score computed with Equation~\ref{eq:gap}.  $G_{IL}$ measures the performance gap between each algorithm and an upper bound method.
This upper-bound is represented by $Full$ a non-incremental learning with all data available. 

\begin{equation}
       G_{IL} = \frac{1}{T} \times \sum_{t=1}^{T} \frac{acc_t - acc_{Full}}{acc_{Max} - acc_{Full}} 
    \label{eq:gap}    
\end{equation}

where: $T$ - number of tested configurations; $acc_{t}$ - top-5 score for each configuration (individual values of each row of Table~\ref{tab:big}); $acc_{Full}$ - the upper-bound accuracy of the dataset ($Full$ in Table~\ref{tab:big}); $acc_{Max}$ - the maximum theoretical value obtainable for the measure ($acc_{Max} = 100$ here). 

$G_{IL}$ estimates the average behavior of each algorithm with respect to the upper bound. 
The denominator is introduced to avoid a disproportionate influence of individual datasets in the aggregate score. 
$G_{IL}$ is necessarily a negative number and the closer its value to zero, the better the method is. 
An ideal method, which reaches the upper bound value in all configurations, gives $G_{IL}=0$.
More details about $G_{IL}$ are discussed in the supplementary material.

\textbf{Baselines.} 
Experiments have been conducted with strong baselines which are either inspired from existing IL literature or introduced here because relevant to $ScaIL$:
\begin{itemize}[leftmargin=*]
\itemsep0.1em
    \item $iCaRL$~\cite{DBLP:conf/cvpr/RebuffiKSL17} - method using fine tuning with classification and distillation losses to prevent catastrophic forgetting and NEM classification to counter data imbalance. 
    \item $BiC$~\cite{DBLP:conf/cvpr/WuCWYLGF19} - introduces a bias correction layer to address the imbalance responsible for catastrophic forgetting. 
    
    \item{$DeeSIL$}~\cite{deesil} - fixed representation IL method which freezes the network after the initial non-incremental state and trains an SVM per class. While similar to the fixed representation from~\cite{DBLP:conf/cvpr/RebuffiKSL17}, an important difference concerns the fact that in~\cite{DBLP:conf/cvpr/RebuffiKSL17} past classifiers are retrained only with exemplars in each IL state. Instead, as allowed by the frozen network, SVMs are trained in the initial state of each class and then reused.  

    \item $FT$ - vanilla fine tuning.
    Unlike existing IL algorithms which use distillation~\cite{DBLP:conf/eccv/CastroMGSA18,DBLP:conf/bmvc/He0SC18,DBLP:journals/corr/abs-1807-02802,DBLP:journals/corr/abs-1902-00829,DBLP:journals/corr/abs-1802-07569,DBLP:journals/corr/abs-1904-01769}, only classification loss is used.
    States are initialized with weights from previous model and all the network layers are allowed to evolve. 
    Exemplars are selected randomly.
   $FT$ is the backbone for all subsequent baselines and for $ScaIL$. 

    \item $FT^{NEM}$ - version of $FT$ in which the classification is done using exemplars instead of the outputs of the deep model. $FT^{NEM}$ is equivalent to a version of $iCaRL$ in which the distillation loss component is ablated.
    
    \item $FT^{BAL}$ - $FT$ followed by a balanced fine tuning proposed by~\cite{DBLP:conf/eccv/CastroMGSA18} to reduce the effect of imbalance. $FT^{BAL}$ is equivalent to a version of end-to-end IL ~\cite{DBLP:conf/eccv/CastroMGSA18} in which the distillation loss component is ablated.
    
    \item{$FT^{L2}$} - adds an L2-normalization layer to the raw classifier weights $\mathcal{C}^{N_k}$ given by model $\mathcal{M}^k$ to reduce bias in favor of new classes in current state $\mathcal{S}^k$.
    
    \item{$FT_{init}$} - the initial classifiers $C^j_i$ of each past class replace the classifiers learned only with the past classes exemplars in $\mathcal{S}^k$. No transformation is applied to $C^j_i$. This is an ablation of the mean-related statistics from $ScaIL$.
    
    \item{$FT_{init}^{L2}$} - version of $FT_{init}$ in which all classifiers are $L2$-normalized to make them more comparable.
    
\end{itemize}
\vspace{0.5em}

\textbf{Implementation.}
ResNet-18 is used as backbone architecture for all methods. 
For $iCaRL$ and $BiC$, we use the public TensorFlow~\cite{DBLP:journals/corr/AbadiBCCDDDGIIK16} implementations provided by authors with their hyperparameters. 
$FT$ and its variants are implemented in PyTorch~\cite{paszke2017automatic}. 
The choice of hyperparameters is largely inspired by the original paper of ResNet-18~\cite{DBLP:conf/cvpr/HeZRS16} and by end-to-end incremental learning~\cite{DBLP:conf/eccv/CastroMGSA18}.  
To discard a potential influence of the deep learning framework, we trained $FT$ for one ILSVRC configuration with $\mathcal{Z}=10$ and $\mathcal{B}=0.5\%$ using Tensorflow. The obtained performance is similar to that reported with Pytorch.
$DeeSIL$ SVMs are implemented using scikit-learn~\cite{DBLP:journals/corr/abs-1201-0490} and their parameters are optimized on the training data since it is hard to hold out validation data in IL.
More implementation details are provided in the supplementary material. 

\begin{table*}[t]
\begin{center}
\resizebox{\textwidth}{!}{
\begin{tabular}{|c|c|c|c|c|c|c|c|c|c|c|c|c|}
\hline
States & \multicolumn{12}{c|}{$\mathcal{Z}=10$}\\
\hline
Dataset & \multicolumn{3}{c|}{ILSVRC} &\multicolumn{3}{c|}{VGGFace2} & \multicolumn{3}{c|}{Landmarks} & \multicolumn{3}{c|}{CIFAR-100} \\
\hline
$\mathcal{B}$ & $2\%$ & $1\%$ & $0.5\%$ & $2\%$ & $1\%$ & $0.5\%$ & $2\%$ & $1\%$ & $0.5\%$ & $2\%$ & $1\%$ & $0.5\%$  \\
\hline
{\small $iCaRL^{herd}$} & 62.5 & 61.4 & 60.9 &  83.9 & 81.4 & 78.2 & 82.5 & 80.5 & 76.2 & 85.1 & 83.7 & 83.2 \\
{\small $BiC^{herd}$} & \textbf{85.5} & \textbf{82.8} & \textbf{79.7} & \textbf{97.3} & \textbf{96.6} & \textbf{95.7} & \textbf{97.9} & \textbf{97.3} & \textbf{96.6} & \textbf{88.8} & \textbf{87.6} & \textbf{83.5} \\
{\small $DeeSIL$}& 74.5 & 74.3 & 74.2 & 92.6 & 92.5 & 92.2& 93.9 & 93.6 & 92.9 & 66.5 & 65.2 & 63.7  \\
{\small $FT$} & 77.0 & 70.1 & 60.0& 96.0 & 94.1 & 90.7& 95.8 & 93.2 & 89.1 & 80.0 & 73.7 & 63.3\\
{\small $FT^{NEM}$ }  & 79.4 & 74.5 & 69.6& 95.7 & 94.1 & 91.0 & 95.2 & 92.7 & 88.8 & 82.4 & 77.4 & 68.4 \\
{\small $FT^{BAL}$}& 81.3 & 78.0 & 72.3 & 96.4 & 95.0 & 92.2& 96.3 & 94.3 & 90.0 & 73.0 & 65.0 & 56.1 \\
{\small $FT^{L2}$ }& 81.4 & 77.6 & 72.1& 96.5 & 95.1 & 92.4 & 96.2 & 94.4 & 91.4 & 81.8 & 77.2 &  69.1  \\

{\small $FT_{init}$ }& 68.9& 66.5 & 61.2 & 95.9 & 95.3 & 94.5 & 96.5 & 95.0 & 92.7 & 79.3 & 77.3 & 73.7 \\
{\small $FT^{L2}_{init}$ }& 78.4 & 75.7 & 73.3& 95.9 & 95.3 & 94.5 & 96.5 & 95.0 & 92.7 &  83.0 & 79.2 & 72.7\\
{\small $ScaIL$ } & 81.0 & 78.2 & 75.1 & 96.4 & 95.6 & 94.5 & 96.9 & 95.3 & 92.7 &  84.6 & 81.1 & 74.9 \\
{\small $ScaIL^{herd}$} & 82.0 & 79.8 & 76.6 & 96.5 & 95.8 & 95.2 & 97.3 & 96.0 & 94.0 & 85.6 & 83.2 & 79.1 \\
\hline 		
$Full$ & \multicolumn{3}{c|}{92.3} &\multicolumn{3}{c|}{99.2} & \multicolumn{3}{c|}{99.1} & \multicolumn{3}{c|}{91.2} \\
\hline
\end{tabular}

\begin{tabular}{|cc|cc|cc|cc|}
\hline
\multicolumn{8}{|c|}{$\mathcal{B} = 0.5\%$}\\
\hline
\multicolumn{2}{|c|}{ILSVRC} & \multicolumn{2}{c|}{VGGFace2} & \multicolumn{2}{c|}{Landmarks}& \multicolumn{2}{c|}{CIFAR-100}  \\
\hline
Z=20  & Z=50  & Z=20 & Z=50 & Z=20 & Z=50 & Z=20 & Z=50 \\
\hline
56.2 & 42.9 & 72.7 & 52.3 & 72.4 & 54.2 & \textbf{73.2} & \textbf{55.7}\\
74.6 & 63.9 & 92.3 & 85.3 & \textbf{94.7} & \textbf{90.5} & 50.5 & 19.6\\
69.0 & 58.0 &  87.2 & 78.9 &  90.6 & 84.8 & 63.4 & 42.5 \\
64.5 & 59.2 & 90.8 & 86.5 & 87.8 & 85.5 &  59.9 & 49.4 \\
72.7 & 63.4 & 91.8 & 87.8 & 88.1 & 86.0 & 64.5 & 51.0 \\
70.5 & 61.1 & 91.7 & 86.5 & 87.8 & 85.3 & 57.1 & 50.0\\
73.4 & 66.7& 92.8 & 88.8 & 89.8 & 87.1 & 63.2 & 49.9 \\
53.4 & 39.0 &  \textbf{95.1} & 90.3 & 90.6 & 87.5  & 60.7 & 40.1 \\
72.0 & 66.0 & \textbf{95.1} & 90.2 & 90.7 & 87.6  & 64.3 & 42.5 \\
73.9 & 68.3 & 94.5 & 90.5 & 90.7 & 88.2 & 67.9 & 47.7\\
\textbf{76.6} & \textbf{70.9} & 95.0 & \textbf{92.4} & 92.6 & 90.4 & 69.8 & 51.0\\
\hline
\multicolumn{2}{|c|}{92.3} &\multicolumn{2}{c|}{99.2} & \multicolumn{2}{c|}{99.1} & \multicolumn{2}{c|}{91.2} \\
\hline

\end{tabular}

\begin{tabular}{|c|}
\hline
\multirow{3}{*}{$G_{IL}$}\\
\\
\\
\hline
-16.75\\
-4.03\\
-7.10\\
-6.40\\
-6.01\\
-5.98\\
-5.17\\
-5.23\\
-4.67\\
-4.41\\
\textbf{-3.71}\\
\hline
-\\
\hline

\end{tabular}

}
\end{center}
\vspace{-1em}
	\caption{Top-5 average accuracy (\%). Following~\cite{DBLP:conf/eccv/CastroMGSA18}, accuracy is averaged only for incremental states (i.e. excluding the initial, non-incremental state). The sizes of past memory $\mathcal{B}$ and number of IL states are varied to evaluate the robustness of algorithms. $Full$ is the non-incremental upper-bound performance obtained with all data available. The methods whose names include $herd$ exploit herding while the others are based on random exemplar selection. \textit{Best results are in bold}.
	}
\label{tab:big}
\end{table*}

\subsection{Discussion of results}
Confirming the conclusions of~\cite{DBLP:conf/cvpr/RebuffiKSL17}, $iCaRL$ has the best overall performance for CIFAR-100 in Table~\ref{tab:big}. 
For the three larger datasets, the $FT$ consistently outperforms $iCaRL$. 
Overall, $FT$ more than halves the gap with $Full$ compared to $iCaRL$ ($G_{IL} = -6.40$ vs. $G_{IL} = -16.75$). 
The comparison to end-to-end IL~\cite{DBLP:conf/eccv/CastroMGSA18}, which achieves 69.4\% top-5 accuracy for ILSVRC with $\mathcal{B}=2\%$ is equally favorable to $FT$\footnote{Note that a complete set of results is not presented for end-to-end IL~\cite{DBLP:conf/eccv/CastroMGSA18}. 
This method was not fully tested because we were not able to reproduce the results presented by the authors since the original implementation is based on Matlab, a non-free environment to which we don't have access.}.
Since one important difference between $FT$ and existing IL methods is the use of distillation, we analyze its role separately in Subsection~\ref{subsec:distill}.

The $FT$-based methods all have a positive contribution. 
$FT^{NEM}$ and $FT^{BAL}$ which are inspired by $iCaRL$~\cite{DBLP:conf/cvpr/RebuffiKSL17} and end-to-end IL~\cite{DBLP:conf/eccv/CastroMGSA18} improve over $FT$ by less than $0.5$ $G_{IL}$ points.
$FT^{L2}$, the L2-normalized version of the classifiers from the current IL state, provides a gain of $1.23$ $G_{IL}$ points compared to $FT$. 
Somewhat surprisingly, the direct concatenation of initial classifier weights from different states in $FT_{init}$ also improves performance over $FT$ by over $1$ point. 
However, its performance for individual configurations is much more contrasted than that of $FT^{L2}$.
$FT_{init}$ has low results for the two object recognition datasets, which are on average more difficult than face and landmark recognition tasks.
$FT^{L2}_{init}$ adds $L2$-normalization to $FT_{init}$ classifiers and ranks fourth among all methods tested, with $1.73$ $G_{IL}$ improvement over $FT$.
The best overall result is obtained with $ScaIL^{herd}$, which improves $FT$ performance by $2.69$ points. 
The difference between $ScaIL$ and $FT^{L2}_{init}$ in terms of $G_{IL}$ is not large but still interesting.
$ScaIL$ has the most stable behavior among all those tested. In fact, its performance on the three large datasets is most interesting for the smallest $\mathcal{B}$ values.
This is the most challenging case and also the most interesting in practice since it requires a reduced memory for past data.
The increase of the number of incremental state results in a drop of performance for all methods. 
With equal memory $\mathcal{B}$, the worst results are obtained for $\mathcal{Z}=50$ states, followed by $\mathcal{Z}=20$ and $\mathcal{Z}=10$.
This finding confirms the results reported in~\cite{DBLP:conf/eccv/CastroMGSA18} and~\cite{DBLP:conf/cvpr/RebuffiKSL17}.
It is probably an effect of a larger number of incremental rehearsal steps which are applied for larger $\mathcal{Z}$. 
Again, $ScaIL$ is the method which is the least affected by the change of the number of incremental states. 

Contrarily to the conclusion of~\cite{DBLP:conf/eccv/CastroMGSA18}, the herding mechanism in $ScaIL^{herd}$ has positive effect compared to random selection of exemplars in $ScaIL$.
Results show that, while $BiC$~\cite{DBLP:conf/cvpr/WuCWYLGF19} is better for a lower number of incremental states ($\mathcal{Z}=10$), $ScaIL$ has better behavior for a larger number of states.
Equally important, $ScaIL$ performance is less affected by the reduction of the memory size and its performance is globally better for $B=0.5\%$, this leads to a better $G_{IL}$ score for $ScaIL$. Finally, the need of $BiC$ for a validation set to parametrize the bias correction layer makes it nonfunctional if no memory of the past is available. 

The performance gap between $Full$ learning and IL is naturally higher for more complex tasks, such as object recognition, compared to face and landmark recognition.
For the last two tasks, classes have a more coherent visual representation and fewer examples are needed for a comprehensive representation of them. 
In the simplest configurations reported here ($\mathcal{Z}=10$, $\mathcal{B}=2\%$), the best IL algorithms are less than three points behind $Full$ for faces and landmarks. 
For such specialized tasks, incremental learning seems thus applicable in practice without a very significant performance loss.
The situation is different for more complex tasks, such as object recognition, where significant progress is needed before IL algorithms approach the performance of classical learning.

An additional result concerns $DeeSIL$, the fixed representation method. 
Here, it is globally better than $iCaRL$, a finding which is at odds with the results originally reported in~\cite{DBLP:conf/cvpr/RebuffiKSL17}.
The difference is explained by the use of all data for each class, while past class training was unnecessarily restricted to $\mathcal{B}$ exemplars in~\cite{DBLP:conf/cvpr/RebuffiKSL17}.
$FT$ outperforms $DeeSIL$ by less than $1$ $G_{IL}$ point.
For $\mathcal{Z}=10$, $DeeSIL$ has very low dependence on the bounded memory size and could be also used in absence of past exemplars memory.
Naturally, its performance drops for larger $\mathcal{Z}$ values because the initial model is learned with fewer classes but remains interesting. 

\subsection{Effect of distillation in IL}
\label{subsec:distill}
The use of knowledge distillation in incremental learning with bounded memory was pioneered in iCaRL~\cite{DBLP:conf/cvpr/RebuffiKSL17}, which extends the work on IL without memory from~\cite{DBLP:conf/cvpr/HeZRS16}. 
Distillation was largely adopted afterwards~\cite{DBLP:conf/eccv/CastroMGSA18,DBLP:conf/bmvc/He0SC18,DBLP:journals/corr/abs-1807-02802,DBLP:journals/corr/abs-1902-00829,DBLP:journals/corr/abs-1802-07569,DBLP:journals/corr/abs-1904-01769} as a way to reduce the effect of catastrophic forgetting. 
This adoption was based on one experiment presented in~\cite{DBLP:conf/cvpr/RebuffiKSL17} which compared the performance of iCaRL and fine tuning only on the CIFAR-100 dataset and with a single memory size.
In Table~\ref{tab:big}, we report a similar finding for this dataset.
For CIFAR-100, $FT$ is probably less effective because it uses hard targets for loss minimization.
These targets encode very sparse information for the small dataset available.
In contrast, distillation exploits soft targets which encode more information~\cite{DBLP:journals/corr/HintonVD15} and is thus more fitted to work with small datasets. 
The results for $\mathcal{Z}=10$ states with different values of $\mathcal{B}$ support the above observation since the difference in favor of $iCaRL$ grows as $\mathcal{B}$ is reduced.

However, distillation hurts performance for all configurations tested for the three larger datasets, where $FT$ has consequently better performance than $iCaRL$.
The use of network outputs as soft targets for distillation was noted to produce a classification bias for past classes both in the original knowledge distillation paper~\cite{DBLP:journals/corr/HintonVD15} and in an incremental context~\cite{DBLP:journals/corr/abs-1807-02802}. 
A common assumption of distillation-based IL algorithms, first  made in~\cite{DBLP:conf/cvpr/HeZRS16}, is that the process starts with a powerful pretrained model which is trained on a large and balanced dataset.
Under this condition, the soft targets used by the distillation loss are efficient to transfer knowledge to the next incremental state. 
Our hypothesis is that distillation tends to reinforce the errors due to data imbalance in the previous incremental state.
In practice, if the distillation component is fed with soft targets whose predictions are wrong, it will push the classifier toward wrong classes.
To verify this hypothesis, we present an analysis of correct and erroneous predictions for past and new classes in Table~\ref{tab:errors} for vanilla fine tuning ($FT$) and fine tuning with distillation used as backbone in $iCaRL$ ($FT^{distill}$).
Results are shown only for ILSVRC with $\mathcal{Z}=10$ states and $\mathcal{B}=5000$ exemplars but trends are similar for other configurations.
The bias toward new classes, expressed by $e(p,n)$ errors is similar with and without distillation.
The correct predictions for new classes are also in a comparable range, although lower for $FT^{distill}$.
This indicates that the data imbalance toward new classes has rather comparable effect regardless of the use of distillation. The performance difference between the two methods is due mainly to confusions between past classes expressed by $e(p,p)$. 
They are roughly three times more frequent for $FT^{distill}$ compared to $FT$ in Table~\ref{tab:errors}.
Equally important, while distillation is supposed to preserve accuracy for past classes, it clearly does not since the amount of correctly recognized past examples grows very steadily in $FT^{distill}$.

\begin{table}[t!]\centering
\resizebox{0.48\textwidth}{!}{
	\begin{tabular}{c|c|cccccccccc}
	     & & \multicolumn{9}{c}{\textbf{Incremental states}}\\
		\toprule   & & $\mathcal{S}^1$ & $\mathcal{S}^2$  & $\mathcal{S}^3$ & $\mathcal{S}^4$ & $\mathcal{S}^5$ & $\mathcal{S}^6$ & $\mathcal{S}^7$ & $\mathcal{S}^8$ & $\mathcal{S}^9$ \\    
		\midrule
		 \parbox[t]{2mm}{\multirow{6}{*}{\rotatebox[origin=c]{90}{$FT$}}}   
        &$c(p)$  & 2117 & 2995 & 3415 & 3875 & 3653 & 4451 & 4558 & 5003 & 3119 \\ 
        &$e(p, p)$  & 156 & 450 & 807 & 1363 & 1842 & 2710 & 2626 & 3932 & 2388 \\ 
        &$e(p, n)$  & 2727 & 6555 & 10778 & 14762 & 19505 & 22839 & 27816 & 31065 & 39493  \\
        &$c(n)$  & 4151 & 4322 & 4103 & 4141 & 4267 & 4304 & 4247 & 4378 & 4248 \\
        &$e(n, n)$  & 809 & 638 & 875 & 828 & 716 & 674 & 743 & 595 & 741 \\
        &$e(n, p)$  & 40 & 40 & 22 & 31 & 17 & 22 & 10 & 27 & 11  \\
		 	\midrule
		 \parbox[t]{2mm}{\multirow{6}{*}{\rotatebox[origin=c]{90}{$FT^{distill}$}}}&
            $c(p)$  & 850 & 1008 & 1355 & 1355 & 1195 & 1344 & 1419 & 1543 & 1562 \\
            &$e(p, p)$  & 472 & 1746 & 3700 & 4999 & 6904 & 8246 & 10771 & 13400 & 14556 \\ 
            &$e(p, n)$  & 3678 & 7246 & 9945 & 13646 & 16901 & 20410 & 22810 & 25057 & 28882 \\ 
            &$c(n)$  & 3645 & 3834 & 3597 & 3607 & 3744 & 3754 & 3605 & 3766 & 3662 \\ 
            &$e(n, n)$  & 1043 & 793 & 928 & 905 & 785 & 776 & 828 & 692 & 751 \\ 
            &$e(n, p)$  & 312 & 373 & 475 & 488 & 471 & 470 & 567 & 542 & 587 \\
		\hline 		
	\end{tabular}
}
	\caption{Top-1 ILSVRC correct and wrong classifications for vanilla fine tuning ($FT$), fine tuning with distillation ($FT^{distill}$) with $\mathcal{Z}=10$ and $\mathcal{B}=5000$. 
	$p$ and $n$ stand for past and new classes. $c$ and $e$ indicate correct and erroneous classifications. $e(p,p)$ is to be read as past class examples wrongly predicted as other past classes. $e(p,n)$ is to be read as past class examples wrongly predicted as new classes. Note that top-1 performance is used because the proposed analysis is impossible for top-5 accuracy. 
	}
	\label{tab:top1_conf}
	\label{tab:errors}
\vspace{-4.3mm}
\end{table}

\begin{figure}[ht!]
	\begin{center}
\includegraphics[width=0.45\textwidth,trim={0cm 2cm 0cm 0cm}]{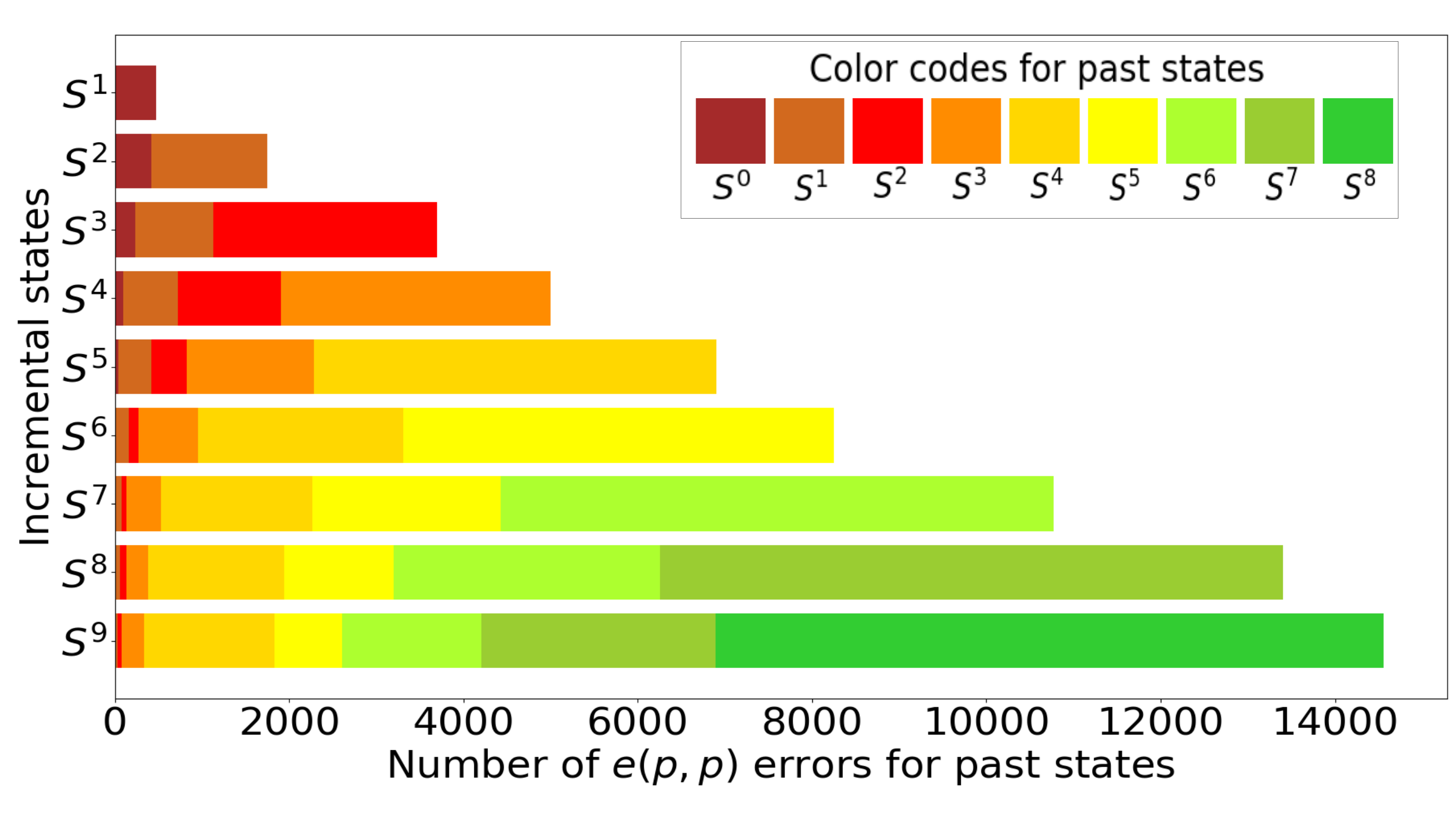}
	\end{center}
	\caption{Detail of past-past errors $e(p,p)$ for individual states of $FT^{distill}$ on ILSVRC with $\mathcal{Z}=10$ and $\mathcal{B}=5000$. We note that, in each state, a majority of errors are due to the latest past state as a result of learning its associated state with an imbalanced training set. \textit{Best viewed in color}.
    }
	\label{fig:err_past}
\end{figure}

In Figure~\ref{fig:err_past}, we present the distribution of $e(p,p)$ errors among individual past states for $FT^{distill}$. 
Since test data is balanced among states, the distribution of errors should also be approximately so. 
Instead, Figure~\ref{fig:err_past} shows that a majority of past test data for state $\mathcal{S}^k$ are predicted as belonging to classes which were new when first learned in $\mathcal{S}^{k-1}$. 
This result confirms that class imbalance has an important role for the distillation component of the loss, similarly to its influence on the classification component.
It is also noticeable that, except for $\mathcal{S}^5$, the number of error grows for more recent past states.
Along with imbalance, the number of rehearsals after the initial learning of the class also plays an important role in terms of distillation-related errors. 

Our findings indicate that vanilla fine tuning is preferable to distillation-based fine tuning as backbone for large scale IL with memory.
Further distillation related experiments are presented in the supplementary material.

\vspace{-1em}

\section{Conclusion}

\vspace{-1em}

We introduced $ScaIL$, a simple but effective IL algorithm which combines classifiers learned in different IL states to reduce catastrophic forgetting.
It keeps the number of parameters of the network constant across IL states and requires a second memory whose size is negligible compared to that of the exemplars memory.
The method is compared to strong state-of-the-art methods, with their improvements based on distillation ablation and with new baselines which exploit initial classifiers.
$ScaIL$ provides performance improvement over published results and is also better than the new baselines. 
Our method is also the most stable over the different memory and IL states values tested. 

A consequent part of the performance improvement is due to the ablation of the distillation  in IL algorithms. 
While widely used, we find that distillation is only useful for small scale datasets.
Our analysis indicates that a performance drop appears for large scale datasets with memory when distillation is used.
The drop is notably due to the inherently imbalanced character of datasets available in IL. 

Comprehensive experiments were run on four public visual datasets with three memory sizes and three numbers of incremental states.
We introduced an aggregated score to get an overview of performance in the different configurations tested.
This experimental protocol can be reused to validate future works. 
To facilitate reproducibility, the code and dataset details are publicly available at: \url{https://github.com/EdenBelouadah/class-incremental-learning}.

The presented results reduce the performance gap between IL algorithms and non-incremental learning but the difference is still important, particularly for harder tasks. 
Class IL with bounded memory remains an open problem and new research is needed to make it usable in practice without significant performance loss.
Future work will aim to: (1) improve vanilla $FT$ while keeping model complexity and memory budget bounded, (2) explore new ways to handle data imbalance and (3) tackle real life situations where streamed data are partially or completely unlabeled.

{\small
\bibliographystyle{ieee}
\bibliography{main}
}

\clearpage
\pagebreak

\part{}
\begin{strip}
\begin{center}

\textbf{\Large Supplementary material for "ScaIL: Classifier Weights Scaling for Class Incremental Learning"}

\vspace{2em}

\author{\large Eden Belouadah, Adrian Popescu\\
Université Paris-Saclay, CEA, Département Intelligence Ambiante et Systèmes Interactifs,\\
91191 Gif-sur-Yvette, France\\
{\tt\small {eden.belouadah,adrian.popescu}@cea.fr}
}
\end{center}
\end{strip}

\section{Introduction}

In this supplementary material, we provide:
\begin{itemize}
    \item a more detailed discussion of $G_{IL}$, the proposed aggregated evaluation score;
    
    \vspace{-0.6em}
    
    \item results for fine tuning with $\mathcal{B}=0$, i.e. without past exemplars memory; 

    \vspace{-0.6em}
    
    \item supplementary experiments related to the role of distillation in class incremental learning;
    
    \vspace{-0.6em}
    
    \item algorithm implementation details.

\end{itemize}

\section{Measuring the performance gap of IL algorithms}
The proposal of aggregated measures is important for tasks which are evaluated in a large number of configurations~\cite{DBLP:conf/nips/RebuffiBV17,tamaazousti2018universal}. 
Building on previous work regarding such measures, the authors of~\cite{tamaazousti2018universal} list eight criteria which should be met by global evaluation metrics when evaluating universal visual representations: (1) coherent aggregation, (2) significance, (3) merit bonus, (4) penalty malus, (5) penalty for damage, (6) independence to outliers, (7) independence to reference and (8) time consistency.
They note that none of the global evaluation measures can fulfill all criteria simultaneously.
However, their formulation which inspired us to propose $G_{IL}$ fulfills the maximum number of criteria. 
While the IL context is different from that of universal representations, a majority of criteria from ~\cite{tamaazousti2018universal} are relevant here. 
The aggregation is easier in our work since the use of $Full$ as reference score is a natural upper bound for incremental learning algorithms. 
The aggregation of scores is natural in $G_{IL}$ since all scores are compared to a single reference. 
The significance criterion, put forward in~\cite{DBLP:conf/nips/RebuffiBV17} is only implicitly modeled because configurations which give the largest gain contribute more to the global score. 
The merit bonus refers to the proportionality of the reward with respect to the reference method and is modeled through the denominator of Equation 3 of the paper. 
The penalty for damage and the penalty malus are not applicable since all methods penalize the performance compared to the upper bound. 
The independence to outlier methods has low effect in our case since it refers to the contributions of individual configurations. 
Since $G_{IL}$ averages the contributions of a relatively large number of contributions, the risk related to outliers is rather reduced. 
Naturally, the more datasets and configurations are tested, the more robust the score will be.
However, the computational resources needed for training in IL are large and we consider that the use of four datasets, with three memory sizes and three incremental learning splits gives a fair idea about the behavior of each algorithm. 
Time consistency is respected since methods are not compared to each other but only to a reference which is stable if the same deep model and data are used across time. 
A question remains whether datasets of different sizes should be given the same weights in the score but using weighting would further complicate the evaluation measure. 

\section{Fine tuning without memory}

\vspace{-1em}

\begin{table}[h]
    \centering 
    \resizebox{0.49\textwidth}{!}{
    \begin{tabular}{|c|c|c|c|c|}
    \hline
        States & \multicolumn{4}{c|}{$\mathcal{Z}=10$} \\
        \hline
        Dataset & ILSVRC & VGGFace2 & Landmarks & CIFAR-100 \\
        \hline
        $LwF$ & 43.80 & 48.30 & 46.34 & 79.49 \\
        $FT^{noMem}$ & 20.64 & 21.28 & 21.29 & 21.27 \\
        $FT^{L2}$ & 20.64 & 21.27 & 21.27 & 21.27 \\
        $FT_{init}$ & 60.95 & 90.90 & 68.77 & 55.05 \\
        $FT^{L2}_{init}$ & 51.57 & 76.84 & 61.42 & 47.48 \\
        $ScaIL$ & 21.96 & 23.06 & 22.31 & 33.49\\
        \hline
    \end{tabular}
}
    \caption{Top-5 accuracy of fine tuning without memory ($\mathcal{B}=0$) for the four datasets with $\mathcal{Z}=10$ states. For reference, we also present $LwF$~\cite{DBLP:conf/cvpr/HeZRS16}, which is equivalent to $iCaRL$~\cite{DBLP:conf/cvpr/RebuffiKSL17} without memory.}
    \label{tab:supp_no_mem_ft_table}
\end{table}

Table~\ref{tab:supp_no_mem_ft_table} provides results obtained with fine tuning without memory for past classes ($\mathcal{B}=0$) and $\mathcal{Z}=10$ states.
Trends are similar for the other $\mathcal{Z}$ values tested in the paper which are not presented here.
The accuracy drops significantly for $FT$ since the network cannot rehearse knowledge related to past classes. Catastrophic forgetting is more severe and past classes become unrecognizable in the current state.  
The accuracy of $FT^{noMem}$ is mostly due to the recognition rate of new classes.
When $\mathcal{Z}=10$, they represent between a half and a tenth of the total number of classes for states $S=1$ and $S=9$, the first and the last incremental state respectively. 
The accuracy for past classes is close to random. 
Since $ScaIL$ depends heavily on the weights of past classes in the current state, its performance drops significantly.
$LwF$~\cite{DBLP:conf/cvpr/HeZRS16} includes a distillation component which is clearly useful in absence of memory.
It outperforms $FT$ and $ScaIL$ for all datasets by a very large margin. 
This finding reinforces the conclusions of~\cite{DBLP:conf/cvpr/RebuffiKSL17} regarding the positive role of distillation in incremental learning without memory.

\begin{figure*}[!htb]
        \centering
        \begin{subfigure}[b, height=4]{0.475\textwidth}
            \centering
            \includegraphics[width=\textwidth]{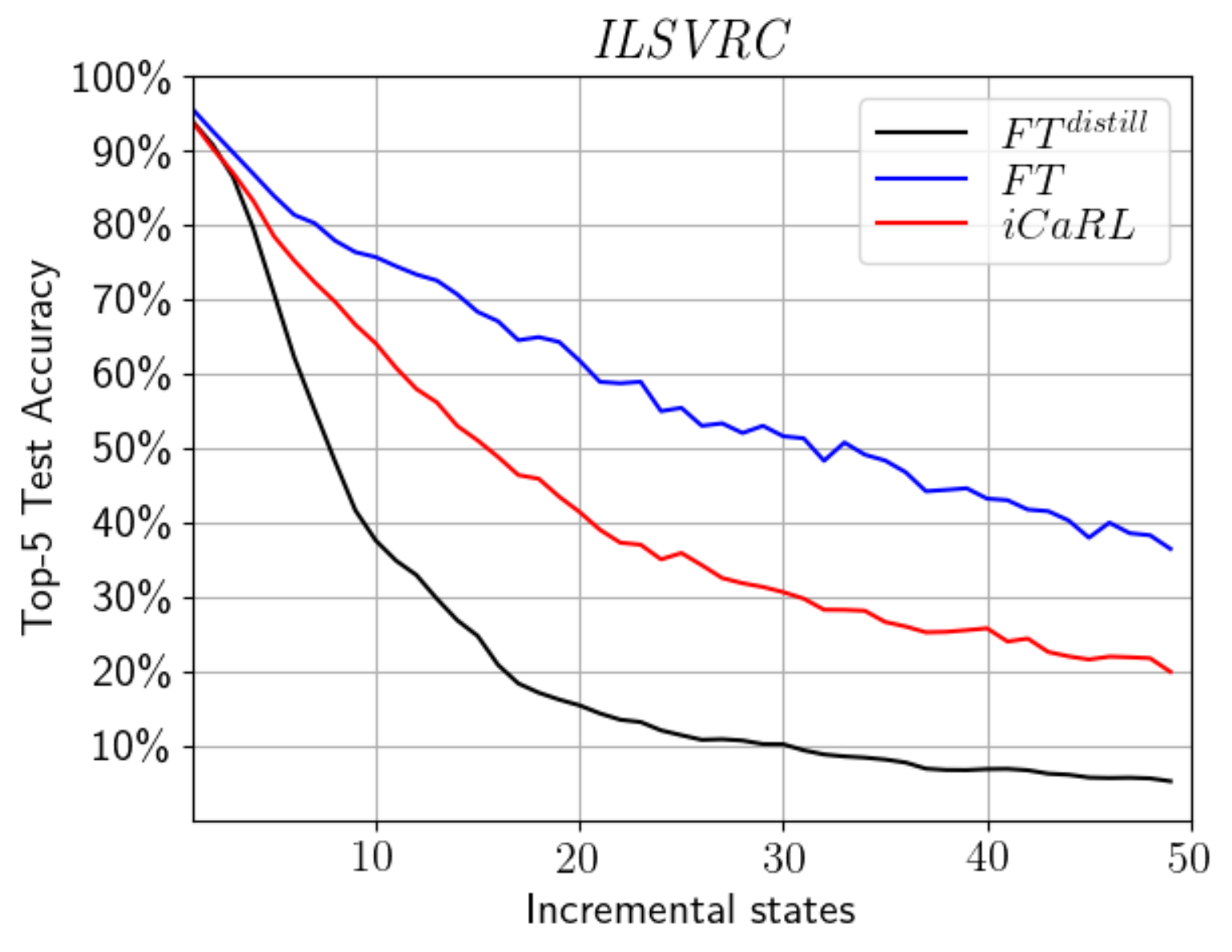}
        \end{subfigure}
        \hfill
        \begin{subfigure}[b, height=4]{0.475\textwidth}  
            \centering 
            \includegraphics[width=\textwidth]{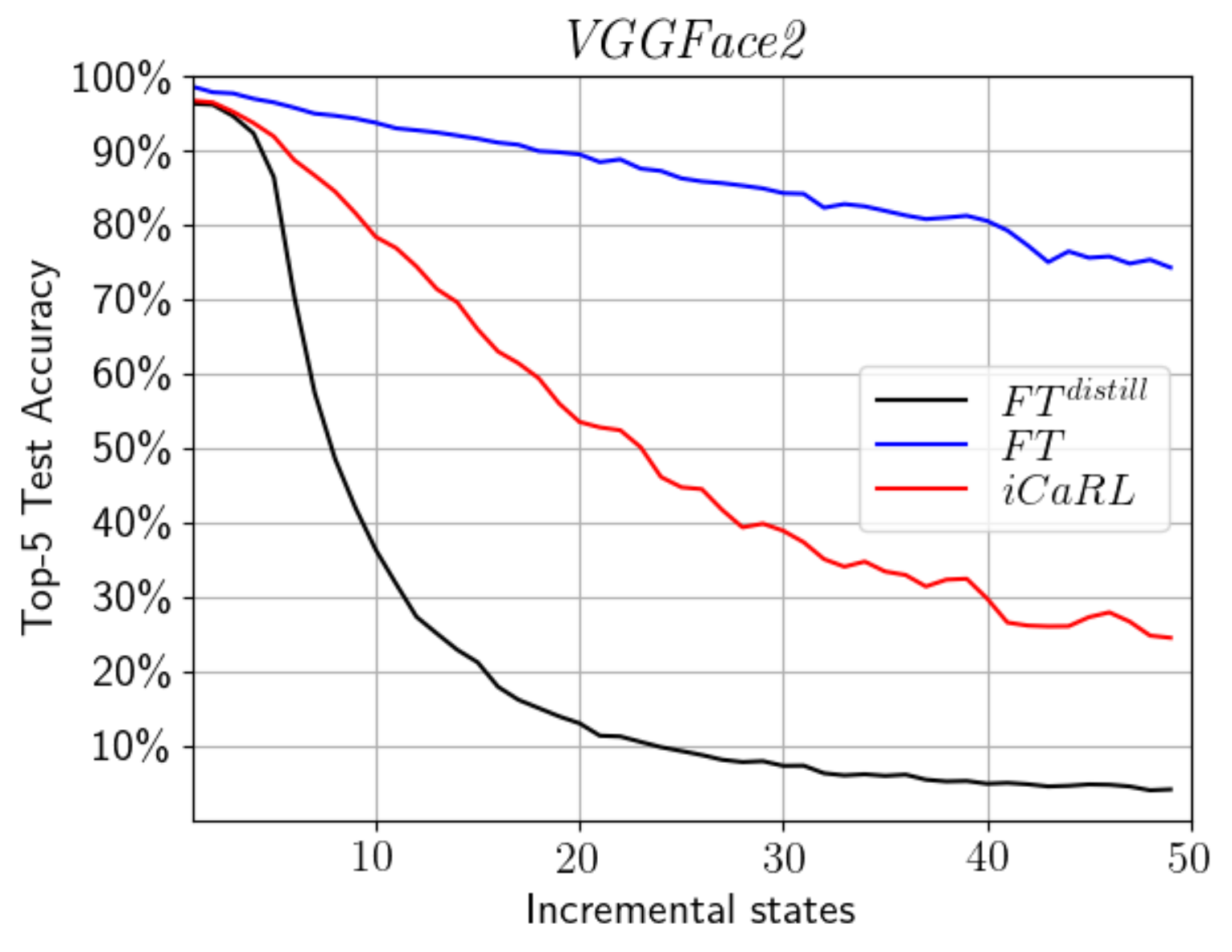}
        \end{subfigure}

        \begin{subfigure}[b, height=4]{0.475\textwidth}   
            \centering 
            \includegraphics[width=\textwidth]{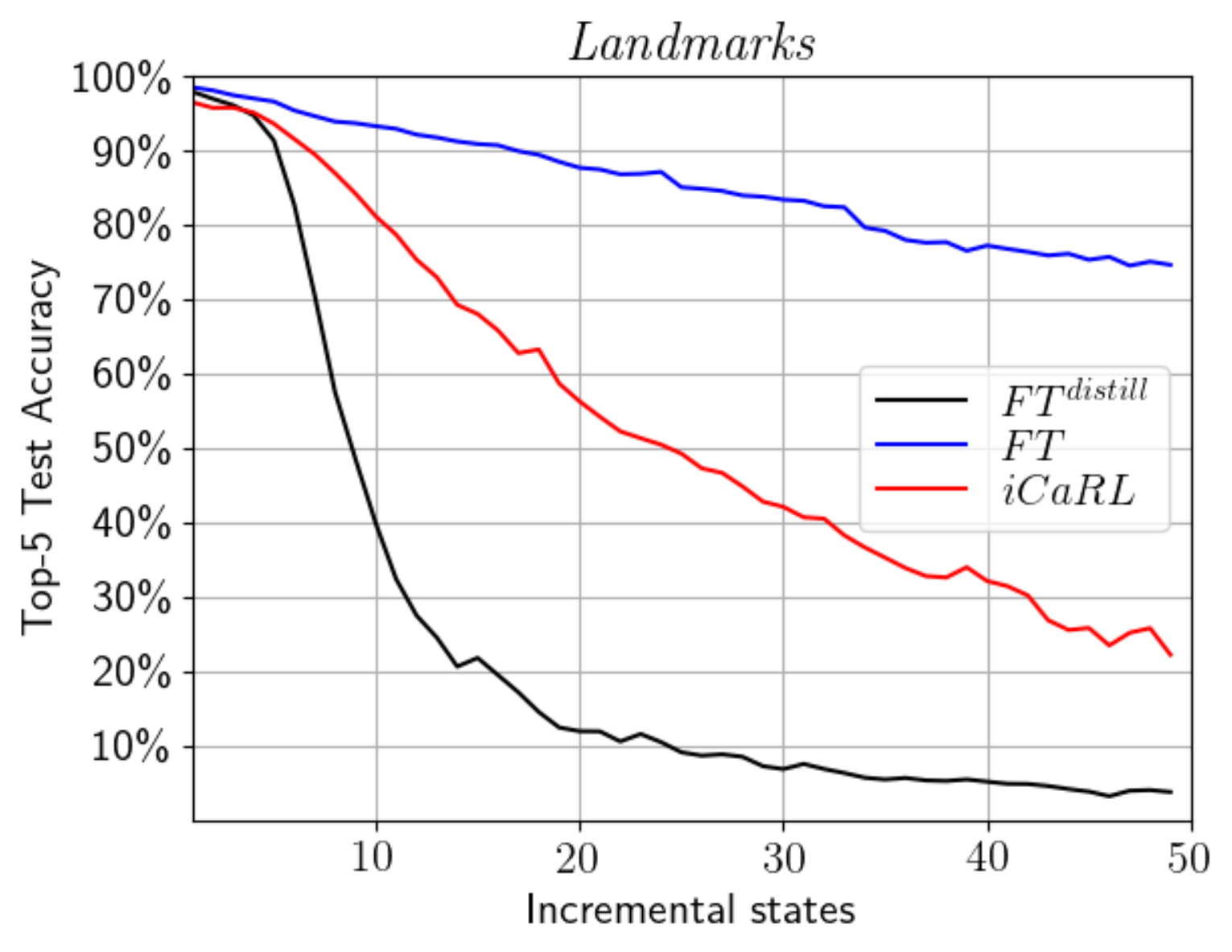}
        \end{subfigure}
        \quad
        \begin{subfigure}[b, height=4]{0.475\textwidth}   
            \centering 
            \includegraphics[width=\textwidth]{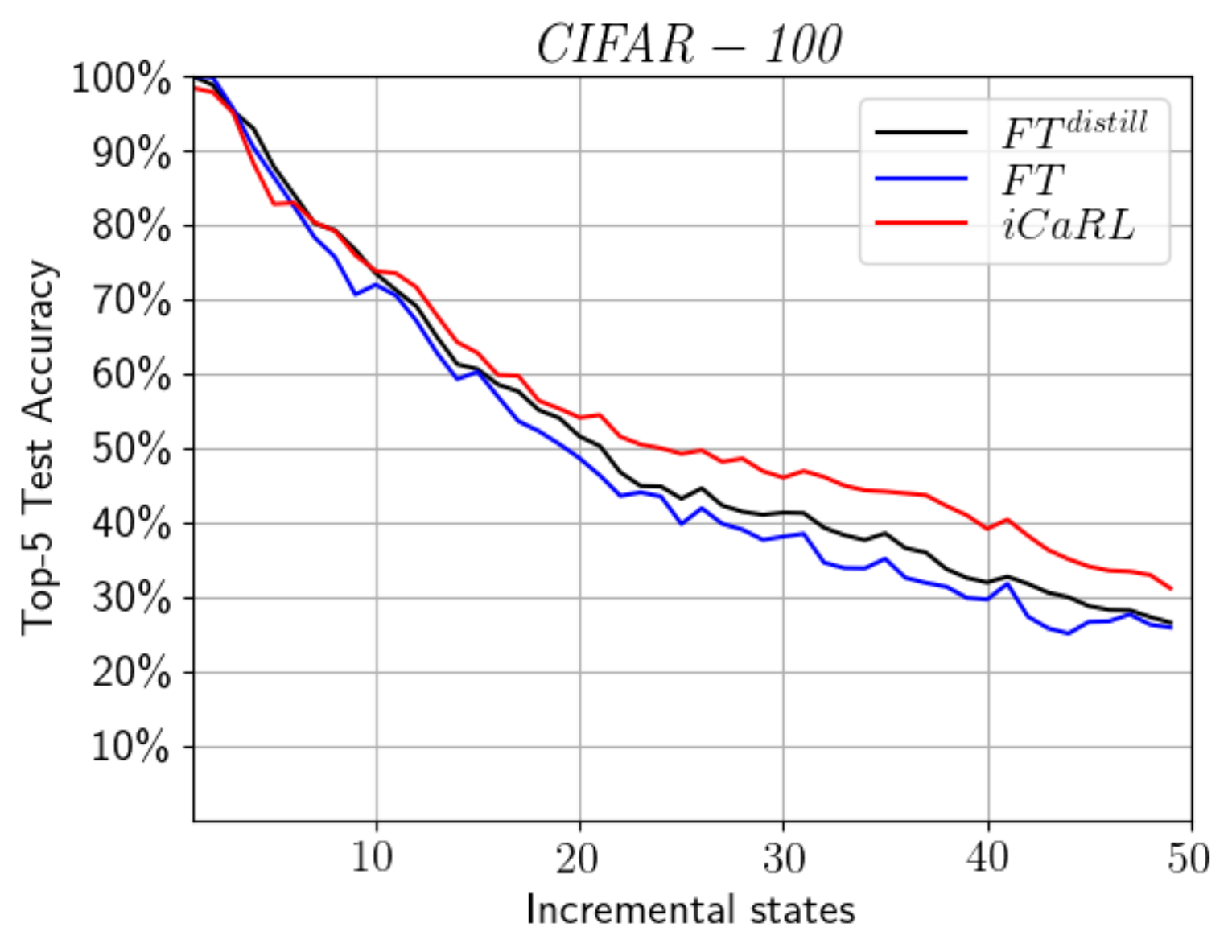}
        \end{subfigure}
        \caption{Detailed Top-5 Test accuracy for the four datasets with $\mathcal{Z}=50$ and memory $\mathcal{B}=0.5\%$. In this experiment, a comparison is done between $FT$, $FT^{distill}$ and $iCaRL$ to analyze the role of distillation.}
        \label{fig:supp_distill_anal}
    \end{figure*}

\begin{figure*}[htb!]

        \centering
        \begin{subfigure}[b]{0.48\textwidth}
            \centering
            ILSVRC
            \includegraphics[width=\textwidth,trim={0cm 1cm 0cm 0cm}]{images/err_bar_ilsvrc_s10_5k.pdf}
        \end{subfigure}
        \hfill
        \begin{subfigure}[b]{0.48\textwidth}  
            \centering 
            VGGFace2
            \includegraphics[width=\textwidth,trim={0cm 1cm 0cm 0cm}]{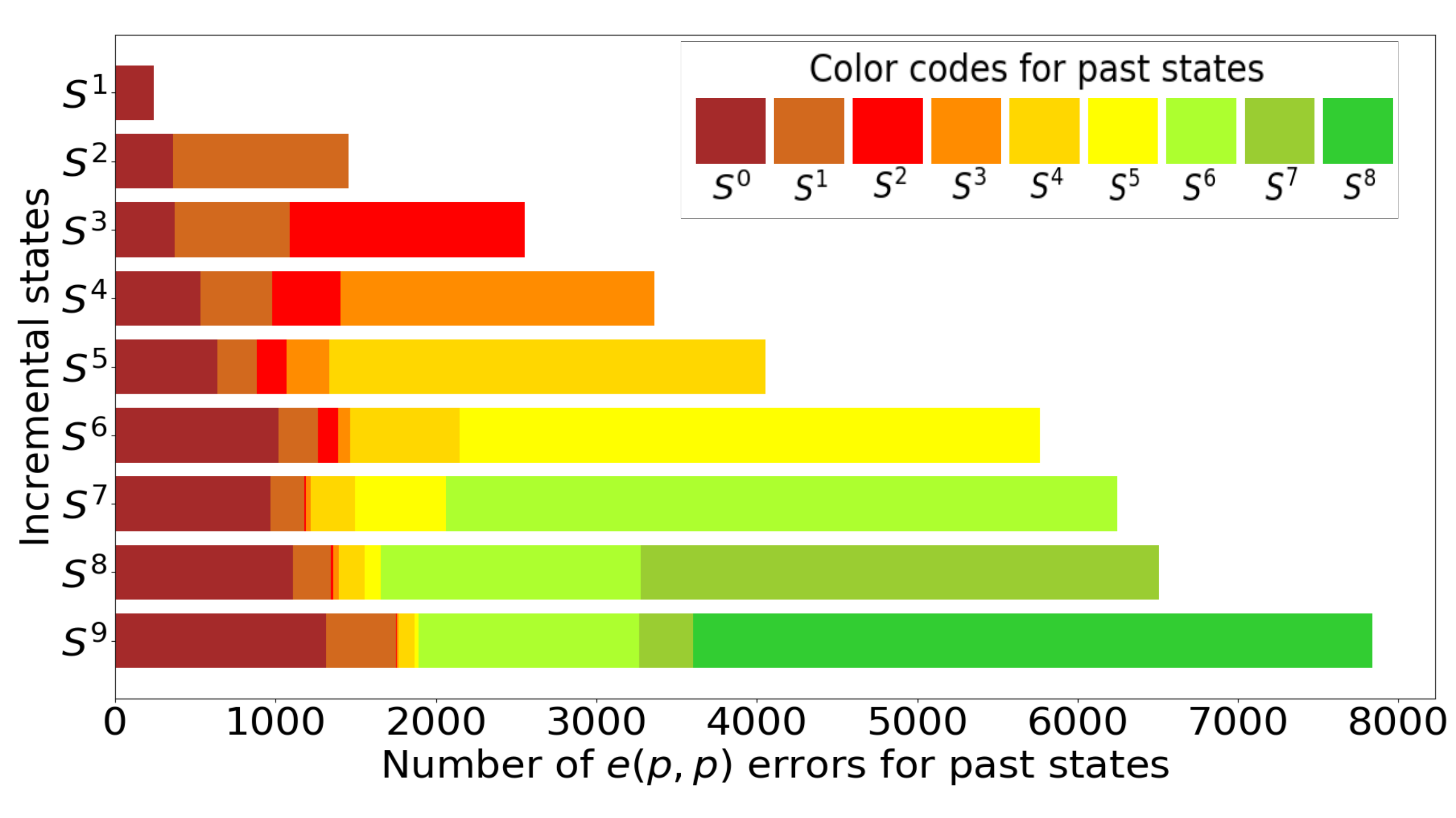}
        \end{subfigure}
        \vskip\baselineskip
        \begin{subfigure}[b]{0.48\textwidth}   
            \centering 
            Landmarks
            \includegraphics[width=\textwidth,trim={0cm 1cm 0cm 0cm}]{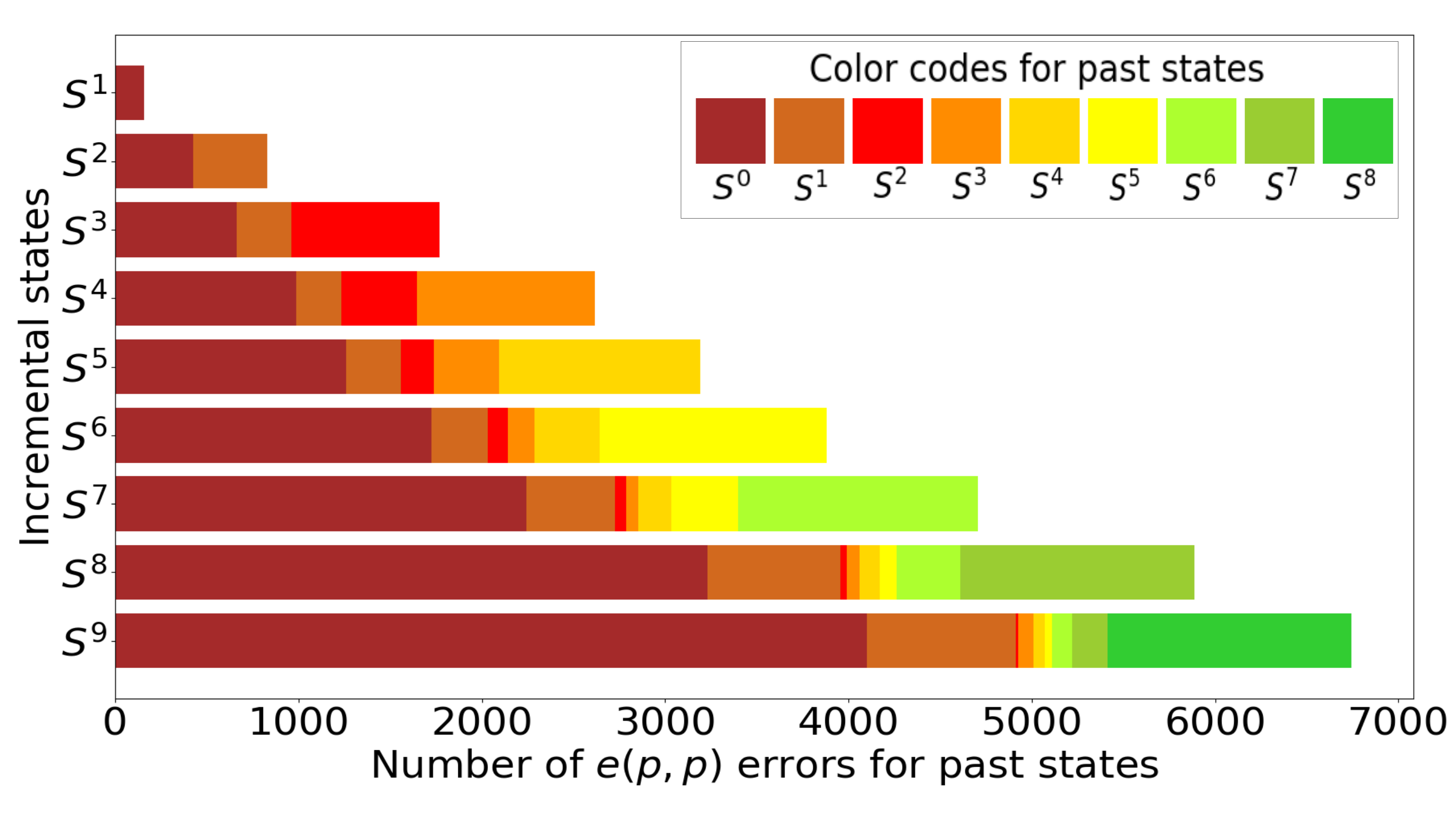}
        \end{subfigure}
        \quad
        \begin{subfigure}[b]{0.48\textwidth}   
            \centering 
            CIFAR-100
            \includegraphics[width=\textwidth,trim={0cm 1cm 0cm 0cm}]{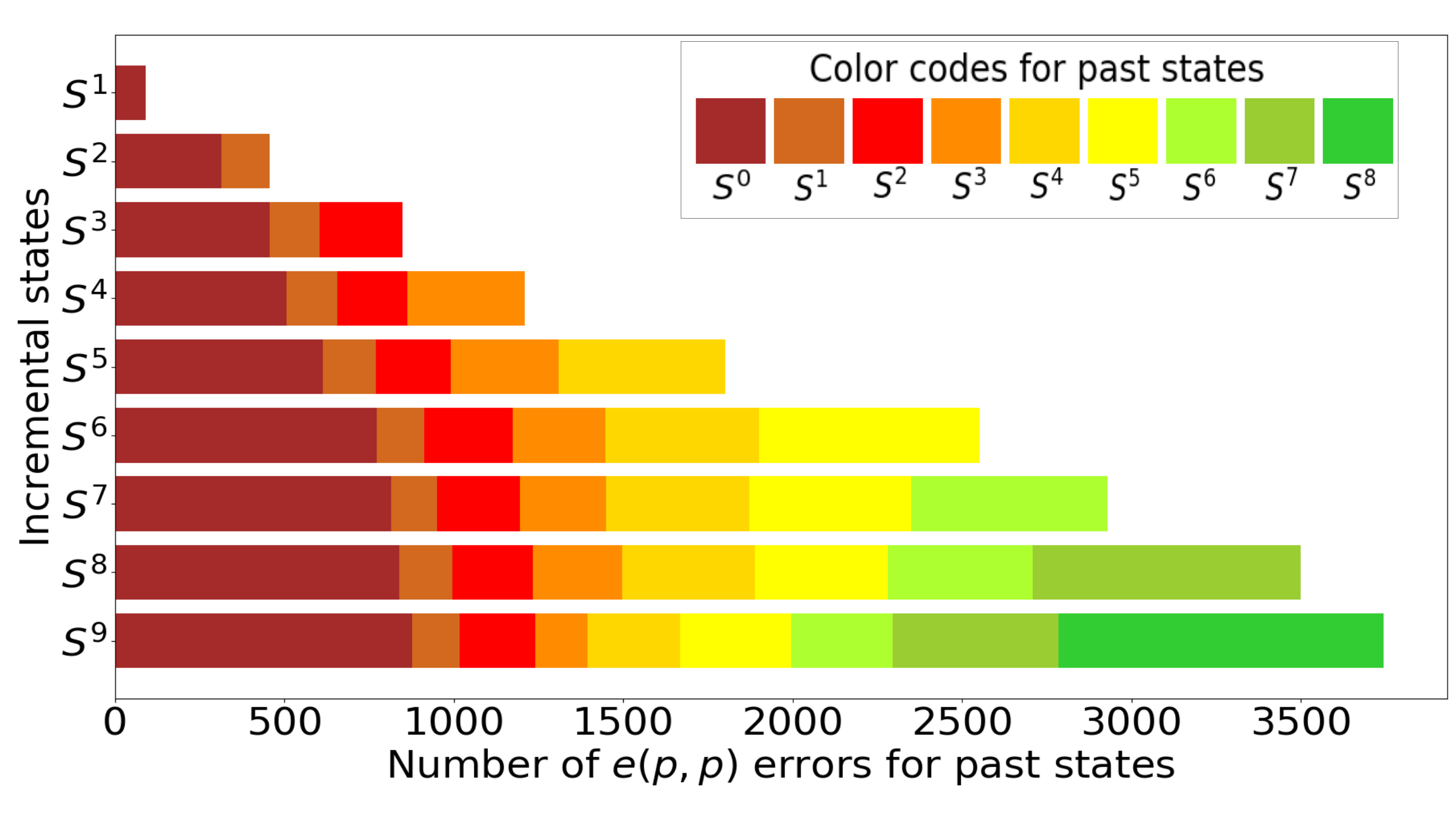}
        \end{subfigure}

	\caption{Detail of past-past errors $e(p,p)$ for  individual states of $FT^{distill}$  on the four datasets with $\mathcal{Z}=10$ and $\mathcal{B}=0.5\%$. In each state, errors due to the latest past state are over-represented as a result of learning its associated state with an imbalanced training set. \textit{Best viewed in color}.
    }
	\label{fig:supp_err_past}
\end{figure*}

\begin{table}[t!]\centering
\resizebox{0.48\textwidth}{!}{
	\begin{tabular}{c|c|cccccccccc}
	     & & \multicolumn{9}{c}{\textbf{Incremental states}}\\
		\toprule   & & $\mathcal{S}^1$ & $\mathcal{S}^2$  & $\mathcal{S}^3$ & $\mathcal{S}^4$ & $\mathcal{S}^5$ & $\mathcal{S}^6$ & $\mathcal{S}^7$ & $\mathcal{S}^8$ & $\mathcal{S}^9$ \\    
		\midrule
		\multicolumn{11}{c}{ILSVRC}\\
		\midrule
		 \parbox[t]{2mm}{\multirow{6}{*}{\rotatebox[origin=c]{90}{$FT$}}}   
		 
		 &$c(p)$  & 2117 & 2995 & 3415 & 3875 & 3653 & 4451 & 4558 & 5003 & 3119 \\ 
        &$e(p, p)$  & 156 & 450 & 807 & 1363 & 1842 & 2710 & 2626 & 3932 & 2388 \\ 
        &$e(p, n)$  & 2727 & 6555 & 10778 & 14762 & 19505 & 22839 & 27816 & 31065 & 39493  \\
        &$c(n)$  & 4151 & 4322 & 4103 & 4141 & 4267 & 4304 & 4247 & 4378 & 4248 \\
        &$e(n, n)$  & 809 & 638 & 875 & 828 & 716 & 674 & 743 & 595 & 741 \\
        &$e(n, p)$  & 40 & 40 & 22 & 31 & 17 & 22 & 10 & 27 & 11  \\
		 	\midrule
		 \parbox[t]{2mm}{\multirow{6}{*}{\rotatebox[origin=c]{90}{$FT^{distill}$}}}&
            $c(p)$  & 850 & 1008 & 1355 & 1355 & 1195 & 1344 & 1419 & 1543 & 1562 \\
            &$e(p, p)$  & 472 & 1746 & 3700 & 4999 & 6904 & 8246 & 10771 & 13400 & 14556 \\ 
            &$e(p, n)$  & 3678 & 7246 & 9945 & 13646 & 16901 & 20410 & 22810 & 25057 & 28882 \\ 
            &$c(n)$  & 3645 & 3834 & 3597 & 3607 & 3744 & 3754 & 3605 & 3766 & 3662 \\ 
            &$e(n, n)$  & 1043 & 793 & 928 & 905 & 785 & 776 & 828 & 692 & 751 \\ 
            &$e(n, p)$  & 312 & 373 & 475 & 488 & 471 & 470 & 567 & 542 & 587 \\
            
		\hline 	
		\multicolumn{11}{c}{VGGFace2}\\
		\midrule
		 \parbox[t]{2mm}{\multirow{6}{*}{\rotatebox[origin=c]{90}{$FT$}}}  
&$c(p)$  & 4168 & 7718 & 11062 & 14293 & 15953 & 19614 & 21075 & 24690 & 24196 \\
&$e(p, p)$  & 89 & 282 & 611 & 947 & 1354 & 2170 & 3203 & 3827 & 4929\\ 
&$e(p, n)$  & 743 & 2000 & 3327 & 4760 & 7693 & 8216 & 10722 & 11483 & 15875\\
&$c(n)$  & 4825 & 4834 & 4866 & 4865 & 4881 & 4879 & 4887 & 4874 & 4883 \\ 
&$e(n, n)$  & 155 & 143 & 118 & 119 & 108 & 102 & 101 & 108 & 108 \\ 
&$e(n, p)$  & 20 & 23 & 16 & 16 & 11 & 19 & 12 & 18 & 9 \\	
\midrule
\parbox[t]{2mm}{\multirow{6}{*}{\rotatebox[origin=c]{90}{$FT^{distill}$}}}   
&$c(p)$  & 1729 & 2109 & 1886 & 1787 & 1520 & 1657 & 1412 & 1199 & 1131 \\
&$e(p, p)$  & 242 & 1455 & 2553 & 3360 & 4056 & 5766 & 6248 & 6506 & 7838 \\ 
&$e(p, n)$  & 3029 & 6436 & 10561 & 14853 & 19424 & 22577 & 27340 & 32295 & 36031 \\ 
&$c(n)$  & 4620 & 4637 & 4694 & 4740 & 4747 & 4714 & 4693 & 4685 & 4728 \\ 
&$e(n, n)$  & 299 & 239 & 236 & 203 & 212 & 224 & 218 & 248 & 216 \\ 
&$e(n, p)$  & 81 & 124 & 70 & 57 & 41 & 62 & 89 & 67 & 56 \\
\hline 	
\multicolumn{11}{c}{Landmarks}\\
\midrule
\parbox[t]{2mm}{\multirow{6}{*}{\rotatebox[origin=c]{90}{$FT$}}}  
&$c(p)$  & 1670 & 3072 & 4476 & 5550 & 6564 & 7626 & 8081 & 9303 & 10309 \\
&$e(p, p)$  & 38 & 131 & 318 & 616 & 879 & 1005 & 1340 & 1961 & 2237 \\ 
&$e(p, n)$  & 292 & 797 & 1206 & 1834 & 2557 & 3369 & 4579 & 4736 & 5454 \\ 
&$c(n)$  & 1945 & 1970 & 1959 & 1956 & 1973 & 1966 & 1975 & 1973 & 1971 \\ 
&$e(n, n)$  & 51 & 27 & 35 & 37 & 24 & 27 & 25 & 23 & 27\\ 
&$e(n, p)$  & 4 & 3 & 6 & 7 & 3 & 7 & 0 & 4 & 2 \\
\midrule
\parbox[t]{2mm}{\multirow{6}{*}{\rotatebox[origin=c]{90}{$FT^{distill}$}}}  
&$c(p)$  & 901 & 1011 & 859 & 815 & 788 & 769 & 622 & 533 & 419  \\ 
&$e(p, p)$  & 159 & 831 & 1770 & 2617 & 3194 & 3880 & 4708 & 5889 & 6744  \\ 
&$e(p, n)$  & 940 & 2158 & 3371 & 4568 & 6018 & 7351 & 8670 & 9578 & 10837  \\ 
&$c(n)$  & 1893 & 1893 & 1902 & 1910 & 1937 & 1913 & 1949 & 1926 & 1936\\ 
&$e(n, n)$  & 66 & 53 & 58 & 61 & 37 & 53 & 36 & 52 & 38  \\
&$e(n, p)$  & 41 & 54 & 40 & 29 & 26 & 34 & 15 & 22 & 26 \\
\midrule
\multicolumn{11}{c}{CIFAR-100}\\
\midrule
\parbox[t]{2mm}{\multirow{6}{*}{\rotatebox[origin=c]{90}{$FT$}}}  
&$c(p)$  & 366 & 614 & 675 & 605 & 686 & 950 & 779 & 692 & 467 \\
&$e(p, p)$  & 10 & 181 & 312 & 288 & 641 & 974 & 835 & 732 & 601  \\
&$e(p, n)$  & 624 & 1205 & 2013 & 3107 & 3673 & 4076 & 5386 & 6576 & 7932  \\
&$c(n)$  & 791 & 873 & 886 & 866 & 848 & 859 & 834 & 888 & 915 \\ 
&$e(n, n)$  & 196 & 114 & 103 & 131 & 146 & 127 & 159 & 104 & 80  \\
&$e(n, p)$  & 13 & 13 & 11 & 3 & 6 & 14 & 7 & 8 & 5  \\
\midrule
\parbox[t]{2mm}{\multirow{6}{*}{\rotatebox[origin=c]{90}{$FT^{distill}$}}}   
&$c(p)$  & 719 & 1160 & 1507 & 1706 & 1988 & 2195 & 2349 & 2404 & 2251 \\
&$e(p, p)$  & 91 & 457 & 847 & 1210 & 1800 & 2551 & 2929 & 3499 & 3743\\ 
&$e(p, n)$  & 190 & 383 & 646 & 1084 & 1212 & 1254 & 1722 & 2097 & 3006\\ 
&$c(n)$  & 694 & 742 & 735 & 752 & 723 & 767 & 708 & 786 & 814 \\ 
&$e(n, n)$  & 78 & 62 & 40 & 53 & 48 & 35 & 57 & 38 & 28 \\ 
&$e(n, p)$  & 228 & 196 & 225 & 195 & 229 & 198 & 235 & 176 & 158 \\	
\midrule
\end{tabular}
}
\caption{Top-1 correct and wrong classifications for vanilla fine tuning ($FT$) and fine tuning with distillation ($FT^{distill}$) for the four datasets with $\mathcal{Z}=10$ and $\mathcal{B}=0.5\%$. 
	}
\label{tab:supp_errors}
\end{table}

\section{Supplementary experiments related to distillation in IL}
In Figure~\ref{fig:supp_distill_anal}, we provide detailed top-5 accuracy per incremental state for $FT$, $FT^{distill}$ and $iCaRL$ for $\mathcal{B}=0.5\%$ and $\mathcal{Z}=50$ states.
The largest value of $Z$ from the paper was chosen in order to observe the behavior with and without distillation for a small number of classes per incremental state. 
For ILSVRC, VGGFace2 and Landmarks, the difference between $FT$ and $FT^{distill}$ is small for initial incremental states, increases a lot afterwards and tends to decrease toward the end of the process but remains very large.
This behavior is explained by the fact that, since past memory is only $\mathcal{B}=0.5\%$, the number of exemplars per class becomes very small toward the end.
For instance, $\mathcal{B}$ includes 5000 images for ILSVRC and there will be only 5 exemplars per class in the last states of the incremental process. 
It is noticeable that rehearsal in $FT$ still works with such a small number of exemplars. 
These finding provides further support to the results reported in the paper regarding the negative role of distillation at large scale for imbalanced datasets when a memory of the past is available. 
Confirming the results from~\cite{DBLP:conf/cvpr/RebuffiKSL17}, distillation is indeed useful for CIFAR-100, where its performance is slightly better than that of $FT$. 
Also, the introduction of an external classifier in $iCaRL$ is clearly useful. 

In Table~\ref{tab:supp_errors} and Figure~\ref{fig:supp_err_past}, we extend the analysis of top-1 types of errors presented in Table 2 and Figure 4 of the paper to the four datasets. 
The $e(p,p)$ errors related to the last incremental state are overrepresented for all four datasets compared. 
However, the errors toward the first incremental state are also better represented for VGGFace2 and even become dominant for Landmarks and CIFAR-100.
This behavior is probably due to the fact that the initial state is stronger for easier tasks.
In these cases, the model evolves to a lesser extent compared to ILSVRC, a more complex visual task. 

\vspace{-1em}

\section{Algorithm implementation details}
We used the Github\footnote{https://github.com/srebuffi/iCaRL} public implementation from~\cite{DBLP:conf/cvpr/RebuffiKSL17} to run $iCaRL$ on TensorFlow~\cite{DBLP:journals/corr/AbadiBCCDDDGIIK16} with the same hyper-parameters and training settings provided by the authors. Hyperparameters are as follows: $lr=2.0$, $weight~decay=0.00001$, $momentum=0.9$, $batch~size =128$. $iCaRL$ was run with a total of 60 epochs for the large datasets and for 70 epochs for CIFAR-100. The learning rate is divided by 5 at $epoch = \{20, 30, 40, 50\}$ for the large datasets and at $epoch = \{49, 63\}$ for CIFAR-100. We tried to optimize the learning process by changing hyperparameters but couldn't improve the results presented by the original authors. 

$BiC$~\cite{DBLP:conf/cvpr/WuCWYLGF19} was also run using the public Github implementation\footnote{https://github.com/wuyuebupt/LargeScaleIncrementalLearning} provided by the authors and the same hyper-parameters.

All the other methods were implemented in Pytorch~\cite{paszke2017automatic} with $batch~size=256$ ($128$ for CIFAR-100), $weight~decay=0.0001$ ($0.0005$ for CIFAR-100) and a $momentum=0.9$. The first non-incremental state was trained for 100 epochs for large datasets and 300 epochs for CIFAR-100. The learning rate is set to 0.1 and divided by 10 when the error plateaus for 10 consecutive epochs (60 epochs for CIFAR-100).
$FT$ was run for 35 epochs (60 epochs for CIFAR-100). The only change compared to the standard training was to set initial learning rate per incremental state at $lr = \frac{0.1}{k+1}$, with $1 \leq k \leq \mathcal{Z}-1$. This results in a gain of less than 1 top-5 accuracy point for ILSVRC with $\mathcal{Z}=10$ and $\mathcal{B}=0.5\%$. During training, the learning rate is divided by 10 when the error plateaus for 5 epochs (15 epochs for CIFAR-100).

The balanced fine tuning performed after $FT$ in $FT^{BAL}$ was run for 15 more epochs (30 epochs for CIFAR-100) and the learning rate is reinitialized to $lr = \frac{0.01}{k+1}$. We also tried to initialize the balanced fine tuning with $lr = \frac{0.1}{k+1}$ and continue from the last learning rate of the imbalanced fine tuning but results were lower. Equally important, training with more epochs did not provide any gain.

The fixed representation in $DeeSIL$~\cite{deesil} is trained only with data from the first incremental batch.
No external data was used to ensure that the method is comparable with the others.
SVM training is done using the \textit{scikit-learn} framework~\cite{DBLP:journals/corr/abs-1201-0490}. 
SVMs were optimized by dividing the IL training set to $\frac{90}{10}$ train/val subsets and iterate through the values of the regularizer $C=\{0.0001, 0.001, 0.01, 1, 10, 100, 1000\}$.
The optimal value was retained for each dataset configuration. 
SVMs are optimized only for the non-incremental state. 
The regularizer is then frozen and used for the subsequent incremental states. We used the default values of the other hyper-parameters provided in $sklearn$.

{\small
\bibliographystyle{ieee}
\bibliography{main}
}

\end{document}